\documentclass[runningheads]{llncs}
\usepackage{comment}
\usepackage{amsmath,amssymb} 
\usepackage{color}

\usepackage{multirow}
\usepackage{caption}
\usepackage{mathtools}
\usepackage{amsfonts}
\usepackage{booktabs}
\usepackage{soul}
\usepackage{color}
\usepackage{xspace}
\usepackage{subcaption}
\usepackage{amsmath}
\usepackage{algorithm}
\usepackage{algpseudocode}
\usepackage{mathtools}
\usepackage{xcolor}
\usepackage{url}
\usepackage{pifont}
\usepackage{hyperref}
\newcommand{\xmark}{\ding{55}}
\usepackage[pdftex]{graphicx}

\newcommand*\samethanks[1][\value{footnote}]{\footnotemark[#1]}

\begin{document}
\pagestyle{headings}
\mainmatter
\def\ECCVSubNumber{1057}  

\title{SCAN: Learning to Classify Images without Labels}
%
\author{Wouter Van Gansbeke$^1$\thanks{Authors contributed equally} \quad Simon Vandenhende$^1$\samethanks[1] \quad Stamatios Georgoulis$^2$\\ \quad Marc~Proesmans$^1$ \quad Luc Van Gool$^{1,2}$}

\institute{$^1$KU Leuven/ESAT-PSI \quad $^2$ETH Zurich/CVL, TRACE}
\authorrunning{W. Van Gansbeke and S. Vandenhende et al.}


\maketitle
\begin{abstract}
Can we automatically group images into semantically meaningful clusters when ground-truth annotations are absent? The task of unsupervised image classification remains an important, and open challenge in computer vision. Several recent approaches have tried to tackle this problem in an end-to-end fashion. In this paper, we deviate from recent works, and advocate a two-step approach where feature learning and clustering are decoupled. First, a self-supervised task from representation learning is employed to obtain semantically meaningful features. Second, we use the obtained features as a prior in a learnable clustering approach. In doing so, we remove the ability for cluster learning to depend on low-level features, which is present in current end-to-end learning approaches. Experimental evaluation shows that we outperform state-of-the-art methods by large margins, in particular $+26.6\%$ on CIFAR10, $+25.0\%$ on CIFAR100-20 and $+21.3\%$ on STL10 in terms of classification accuracy. Furthermore, our method is the first to perform well on a large-scale dataset for image classification. In particular, we obtain promising results on ImageNet, and outperform several semi-supervised learning methods in the low-data regime without the use of any ground-truth annotations. The code is made publicly available \href{www.github.com/wvangansbeke/Unsupervised-Classification.git}{\textbf{here}}.

\end{abstract}

\section{Introduction and prior work}
\label{sec:introduction}
Image classification is the task of assigning a semantic label from a predefined set of classes to an image. For example, an image depicts a cat, a dog, a car, an airplane, etc., or abstracting further an animal, a machine, etc. Nowadays, this task is typically tackled by training convolutional neural networks~\cite{AlexNet,VGG,ResNet,ResNeXt,InceptionNet} on large-scale datasets~\cite{ImageNet,COCO} that contain annotated images, i.e. images with their corresponding semantic label. Under this supervised setup, the networks excel at learning discriminative feature representations that can subsequently be clustered into the predetermined classes. What happens, however, when there is no access to ground-truth semantic labels at training time? Or going further, the semantic classes, or even their total number, are not {\em a priori} known? The desired goal in this case is to group the images into clusters, such that images within the same cluster belong to the same or similar semantic classes, while images in different clusters are semantically dissimilar. Under this setup, unsupervised or self-supervised learning techniques have recently emerged in the literature as an alternative to supervised feature learning. 

\textit{Representation learning} methods~\cite{doersch2015unsupervised,ContextEncoders,zhang2016colorful,noroozi2016unsupervised,RotNet} use self-supervised learning to generate feature representations solely from the images, omitting the need for costly semantic annotations. To achieve this, they use pre-designed tasks, called pretext tasks, which do not require annotated data to learn the weights of a convolutional neural network. Instead, the visual features are learned by minimizing the objective function of the pretext task. Numerous pretext tasks have been explored in the literature, including predicting the patch context~\cite{doersch2015unsupervised,nathan2018improvements}, inpainting patches~\cite{ContextEncoders}, solving jigsaw puzzles~\cite{noroozi2016unsupervised,noroozi2018boosting}, colorizing images~\cite{zhang2016colorful,larsson2017colorization}, using adversarial training~\cite{donahue2017adversarial,donahue2019large}, predicting noise~\cite{bojanowski2017unsupervised}, counting~\cite{noroozi2017representation}, predicting rotations~\cite{RotNet}, spotting artifacts~\cite{jenni2018self}, generating images~\cite{ren2018cross}, using predictive coding~\cite{oord2018representation,henaff2019data}, performing instance discrimination~\cite{wu2018unsupervised,he2019momentum,chen2020simple,CMC,PIRL}, and so on. Despite these efforts, representation learning approaches are mainly used as the first pretraining stage of a two-stage pipeline. The second stage includes fine-tuning the network in a fully-supervised fashion on another task, with as end goal to verify how well the self-supervised features transfer to the new task. When annotations are missing, as is the case in this work, a clustering criterion (e.g. K-means) still needs to be defined and optimized independently. This practice is arguably suboptimal, as it leads to imbalanced clusters~\cite{DeepCluster}, and there is no guarantee that the learned clusters will align with the semantic classes.

As an alternative, \textit{end-to-end~learning} pipelines combine feature learning with clustering. A first group of methods (e.g. DEC~\cite{DEC}, DAC~\cite{DAC}, DeepCluster~\cite{DeepCluster}, DeeperCluster~\cite{DeeperCluster}, or others~\cite{asano20self,haeusser2018associative,yang2016joint}) leverage the architecture of CNNs as a prior to cluster images. Starting from the initial feature representations, the clusters are iteratively refined by deriving the supervisory signal from the most confident samples~\cite{DAC,DEC}, or through cluster re-assignments calculated offline~\cite{DeepCluster,DeeperCluster}. A second group of methods (e.g. IIC~\cite{IIC}, IMSAT~\cite{hu2017learning}) propose to learn a clustering function by maximizing the mutual information between an image and its augmentations.  
In general, methods that rely on the initial feature representations of the network are sensitive to initialization~\cite{DAC,DEC,DeepCluster,DeeperCluster,huang2019unsupervised,haeusser2018associative,yang2016joint}, or prone to degenerate solutions~\cite{DeepCluster,DeeperCluster}, thus requiring special mechanisms (e.g. pretraining, cluster reassignment and feature cleaning) to avoid those situations. Most importantly, since the cluster learning depends on the network initialization, they are likely to latch onto low-level features, like color, which is unwanted for the objective of semantic clustering. To partially alleviate this problem, some works~\cite{IIC,hu2017learning,DeepCluster} are tied to the use of specific preprocessing (e.g. Sobel filtering). 

In this work we advocate a two-step approach for unsupervised image classification, in contrast to recent end-to-end learning approaches. The proposed method, named SCAN (Semantic Clustering by Adopting Nearest neighbors), leverages the advantages of both representation and end-to-end learning approaches, but at the same time it addresses their shortcomings: 
\begin{itemize}
\item In a first step, we learn feature representations through a pretext task. In contrast to representation learning approaches that require K-means clustering after learning the feature representations, which is known to lead to cluster degeneracy~\cite{DeepCluster}, we propose to mine the nearest neighbors of each image based on feature similarity. We empirically found that in most cases these nearest neighbors belong to the same semantic class (see Figure~\ref{fig: neighbors_histogram}), rendering them appropriate for semantic clustering. 
\item In a second step, we integrate the semantically meaningful nearest neighbors as a prior into a learnable approach. We classify each image and its mined neighbors together by using a loss function that maximizes their dot product after softmax, pushing the network to produce both consistent and discriminative (one-hot) predictions. Unlike end-to-end approaches, the learned clusters depend on more meaningful features, rather than on the network architecture. Furthermore, because we encourage invariance w.r.t. the nearest neighbors, and not solely w.r.t. augmentations, we found no need to apply specific preprocessing to the input. 
\end{itemize}

Experimental evaluation shows that our method outperforms prior work by large margins across multiple datasets. Furthermore, we report promising results on the large-scale ImageNet dataset. This validates our assumption that separation between learning (semantically meaningful) features and clustering them is an arguably better approach over recent end-to-end works. 
\section{Method}
\label{sec:method}

The following sections present the cornerstones of our approach. First, we show how mining nearest neighbors from a pretext task can be used as a prior for semantic clustering. Also, we introduce additional constraints for selecting an appropriate pretext task, capable of producing semantically meaningful feature representations. Second, we integrate the obtained prior into a novel loss function to classify each image and its nearest neighbors together. Additionally, we show how to mitigate the problem of noise inherent in the nearest neighbor selection with a self-labeling approach. We believe that each of these contributions are relevant for unsupervised image classification.  

\subsection{Representation learning for semantic clustering}
\label{subsec: method_neighbors}
In the supervised learning setup, each sample can be associated with its correct cluster by using the available ground-truth labels. In particular, the mapping between the images $\mathcal{D} = \left\{X_1, \ldots, X_{|\mathcal{D}|}\right\}$ and the semantic classes $\mathcal{C}$ can generally be learned by minimizing a cross-entropy loss. However, when we do not have access to such ground-truth labels, we need to define a prior to obtain an estimate of which samples are likely to belong together, and which are not. 

End-to-end learning approaches have utilized the architecture of CNNs as a prior~\cite{yang2016joint,DAC,DEC,haeusser2018associative,DeepCluster,DeeperCluster}, or enforced consistency between images and their augmentations~\cite{IIC,hu2017learning} to disentangle the clusters. In both cases, the cluster learning is known to be sensitive to the network initialization. Furthermore, at the beginning of training the network does not extract high-level information from the image yet. As a result, the clusters can easily latch onto low-level features (e.g. color, texture, contrast, etc.), which is suboptimal for semantic clustering. To overcome these limitations, we employ representation learning as a means to obtain a better prior for semantic clustering.

\begin{figure}[t]
\begin{minipage}[t]{.50\linewidth} 
\centering
\includegraphics[width=\textwidth]{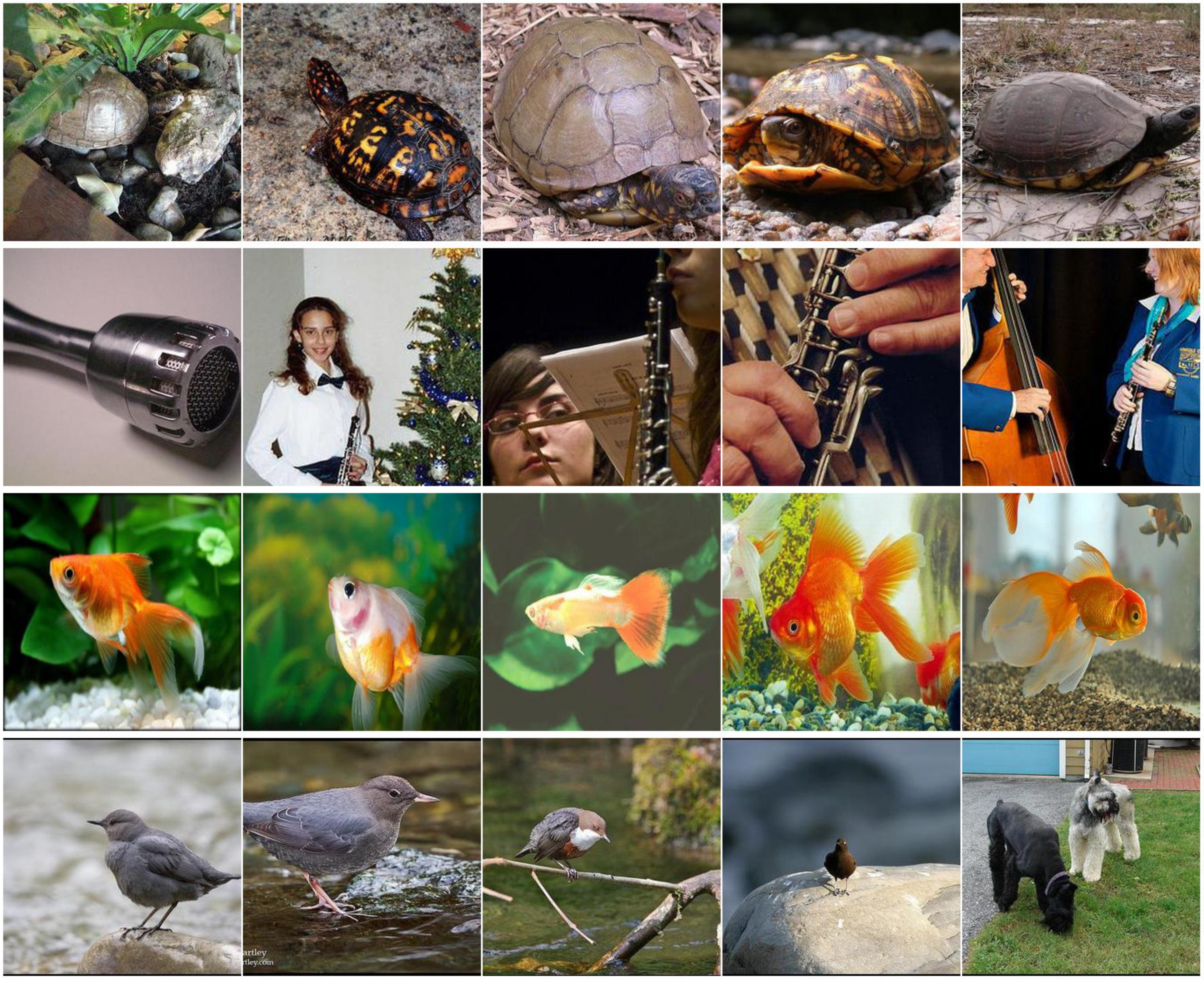}
\caption{Images (first column) and their nearest neighbors (other columns)~\cite{wu2018unsupervised}.}
\label{fig: neighbors_examples}
\end{minipage} %
\hspace*{\fill}
\begin{minipage}[t]{.48\linewidth} 
\centering
\includegraphics[width=\textwidth]{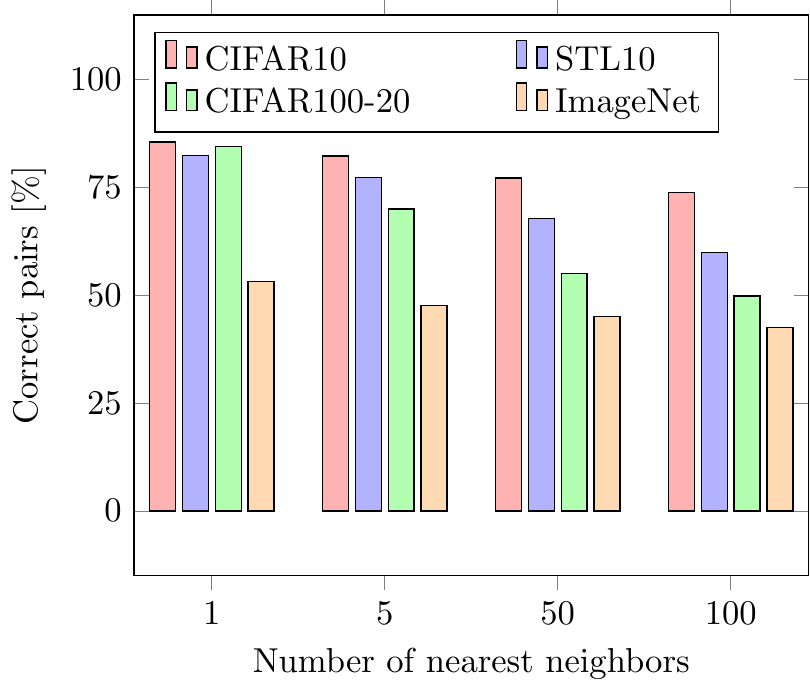}
\caption{Neighboring samples tend to be instances of the same semantic class.}
\label{fig: neighbors_histogram}
\end{minipage} %
\end{figure}

In representation learning, a pretext task $\tau$ learns in a self-supervised fashion an embedding function $\Phi_\theta$ - parameterized by a neural network with weights $\theta$ - that maps images into feature representations. The literature offers several pretext tasks which can be used to learn such an embedding function $\Phi_\theta$ (e.g. rotation prediction~\cite{RotNet}, affine or perspective transformation prediction~\cite{zhang2019aet}, colorization~\cite{larsson2017colorization}, in-painting~\cite{ContextEncoders}, instance discrimination~\cite{wu2018unsupervised,he2019momentum,chen2020simple,PIRL}, etc.). In practice, however, certain pretext tasks are based on specific image transformations, causing the learned feature representations to be covariant to the employed transformation. 
For example, when $\Phi_\theta$ predicts the transformation parameters of an affine transformation, different affine transformations of the same image will result in distinct output predictions for $\Phi_\theta$. 
This renders the learned feature representations less appropriate for semantic clustering, where feature representations ought to be invariant to image transformations. To overcome this issue, we impose the pretext task $\tau$ to also minimize the distance between images $X_i$ and their augmentations $T[X_i]$, which can be expressed as:
\begin{equation}
\label{eq:invariance_objective}
    \min_{\theta}d(\Phi_\theta(X_i), \Phi_\theta(T[X_i])).
\end{equation}
Any pretext task~\cite{wu2018unsupervised,he2019momentum,chen2020simple,PIRL} that satisfies Equation~\ref{eq:invariance_objective} can consequently be used.
For example, Figure~\ref{fig: neighbors_examples} shows the results when retrieving the nearest neighbors under an instance discrimination task~\cite{wu2018unsupervised} which satisfies Equation~\ref{eq:invariance_objective}. We observe that similar features are assigned to semantically similar images. An experimental evaluation using different pretext tasks can be found in Section~\ref{subsec: experiments_ablation}. 

To understand why images with similar high-level features are mapped closer together by $\Phi_\theta$, we make the following observations. First, the pretext task output is conditioned on the image, forcing $\Phi_\theta$ to extract specific information from its input. Second, because $\Phi_\theta$ has a limited capacity, it has to discard information from its input that is not predictive of the high-level pretext task. For example, it is unlikely that $\Phi_\theta$ can solve an instance discrimination task by only encoding color or a single pixel from the input image. As a result, images with similar high-level characteristics will lie closer together in the embedding space of $\Phi_\theta$. 

We conclude that pretext tasks from representation learning can be used to obtain semantically meaningful features. Following this observation, we will leverage the pretext features as a prior for clustering the images. 

\subsection{A semantic clustering loss}
\label{subsec: method_loss}

\noindent\textbf{Mining nearest neighbors.} In Section~\ref{subsec: method_neighbors}, we motivated that a pretext task from representation learning can be used to obtain semantically meaningful features. However, naively applying K-means on the obtained features can lead to cluster degeneracy~\cite{DeepCluster}. A discriminative model can assign all its probability mass to the same cluster when learning the decision boundary. This leads to one cluster dominating the others. Instead, we opt for a better strategy.

Let us first consider the following experiment. Through representation learning, we train a model $\Phi_\theta$ on the unlabeled dataset $\mathcal{D}$ to solve a pretext task $\tau$, i.e. instance discrimination~\cite{chen2020simple,he2019momentum}. Then, for every sample $X_i \in \mathcal{D}$, we mine its $K$ nearest neighbors in the embedding space $\Phi_\theta$. We define the set $\mathcal{N}_{X_i}$ as the neighboring samples of $X_i$ in the dataset $\mathcal{D}$. Figure~\ref{fig: neighbors_histogram} quantifies the degree to which the mined nearest neighbors are instances of the same semantic cluster. We observe that this is largely the case across four datasets\footnote{The details for each dataset are provided in the supplementary materials.} (CIFAR10~\cite{CIFAR}, CIFAR100-20~\cite{CIFAR}, STL10~\cite{STL} and ImageNet~\cite{ImageNet}) for different values of $K$. Motivated by this observation, we propose to adopt the nearest neighbors obtained through the pretext task $\tau$ as our prior for semantic clustering.

\noindent\textbf{Loss function.} We aim to learn a clustering function $\Phi_\eta$ - parameterized by a neural network with weights $\eta$ - that classifies a sample $X_i$ and its mined neighbors $\mathcal{N}_{X_i}$ together. The function $\Phi_\eta$ terminates in a softmax function to perform a soft assignment over the clusters $\mathcal{C}=\left\{1,\ldots,C\right\}$, with $\Phi_\eta \left(X_i\right) \in [0,1]^C$. The probability of sample $X_i$ being assigned to cluster $c$ is denoted as $\Phi_\eta^c(X_i)$. We learn the weights of $\Phi_\eta$ by minimizing the following objective:
\begin{equation}
\label{eq:loss_objective}
\begin{split}
\Lambda = -\frac{1}{|\mathcal{D}|}\sum\limits_{X\in\mathcal{D}}\sum\limits_{k\in\mathcal{N}_{X}}&\log\left<\Phi_\eta(X),\Phi_\eta(k)\right> + \lambda\sum_{c\in\mathcal{C}} \Phi_\eta'^c \log \Phi_\eta'^c, \\
&\text{with~} \Phi_\eta'^c = \frac{1}{|\mathcal{D}|}\sum\limits_{X\in\mathcal{D}}\Phi_\eta^c(X).
\end{split}
\end{equation}

Here, $\left<\cdot\right>$ denotes the dot product operator. The first term in Equation~\ref{eq:loss_objective} imposes $\Phi_\eta$ to make consistent predictions for a sample $X_i$ and its neighboring samples $\mathcal{N}_{X_i}$. Note that, the dot product will be maximal when the predictions are one-hot (confident) and assigned to the same cluster (consistent). To avoid $\Phi_\eta$ from assigning all samples to a single cluster, we include an entropy term (the second term in Equation~\ref{eq:loss_objective}), which spreads the predictions uniformly across the clusters $\mathcal{C}$. If the probability distribution over the clusters $\mathcal{C}$ is known in advance, which is not the case here, this term can be replaced by KL-divergence. 

Remember that, the exact number of clusters in $\mathcal{C}$ is generally unknown. However, similar to prior work~\cite{DEC,DAC,IIC}, we choose $C$ equal to the number of ground-truth clusters for the purpose of evaluation. In practice, it should be possible to obtain a rough estimate of the amount of clusters\footnote{As an example, say you want to cluster various animal species observed in a national park. In this case, we can rely on prior domain knowledge to make an estimate.}. Based on this estimate, we can overcluster to a larger amount of clusters, and enforce the class distribution to be uniform. We refer to Section~\ref{subsec: experiments_overclustering} for a concrete experiment. 

\noindent\textbf{Implementation details}. For the practical implementation of our loss function, we approximate the dataset statistics by sampling batches of sufficiently large size. During training we randomly augment the samples $X_i$ and their neighbors $\mathcal{N}_{X_i}$. For the corner case $K = 0$, only consistency between samples and their augmentations is imposed. We set $K\geq1$ to capture more of the cluster's variance, at the cost of introducing noise, i.e. not all samples and their neighbors belong to the same cluster. Section~\ref{subsec: experiments_ablation} experimentally shows that choosing $K\geq1$ significantly improves the results compared to only enforcing consistency between samples and their augmentations, as in~\cite{IIC,hu2017learning}. 

\noindent\textbf{Discussion}. Unlike~\cite{radford2015unsupervised,kingma2014adam,vincent2010stacked,bengio2007greedy,ng2011sparse,zhao2015stacked,DEC} we do not include a reconstruction criterion into the loss, since this is not explicitly required by our target task. After all, we are only interested in a few bits of information encoded from the input signal, rather than the majority of information that a reconstruction criterion typically requires. It is worth noting that the consistency in our case is enforced at the level of individual samples through the dot product term in the loss, rather than on an approximation of the joint distribution over the classes~\cite{IIC,hu2017learning}. We argue that this choice allows to express the consistency in a more direct way. 

\subsection{Fine-tuning through self-labeling}
\label{subsec: method_selflabeling}
The semantic clustering loss in Section~\ref{subsec: method_loss} imposed consistency between a sample and its neighbors. More specifically, each sample was combined with $K \geq 1$ neighbors, some of which inevitably do not belong to the same semantic cluster. These false positive examples lead to predictions for which the network is less certain. At the same time, we experimentally observed that samples with highly confident predictions ($p_{max} \approx 1$) tend to be classified to the proper cluster. In fact, the highly confident predictions that the network forms during clustering can be regarded as "prototypes" for each class (see Section~\ref{subsec: experiments_imagenet}). Unlike prior work~\cite{DAC,DeepCluster,DEC}, this allows us to select samples based on the confidence of the predictions in a more reliable manner. Hence, we propose a self-labeling approach~\cite{scudder1965probability,mclachlan1975iterative,sohn2020fixmatch} to exploit the already well-classified examples, and correct for mistakes due to noisy nearest neighbors. 

In particular, during training confident samples are selected by thresholding the probability at the output, i.e. $p_{max} >$ threshold. For every confident sample, a pseudo label is obtained by assigning the sample to its predicted cluster. A cross-entropy loss is used to update the weights for the obtained pseudo labels. To avoid overfitting, we calculate the cross-entropy loss on strongly augmented versions of the confident samples. The self-labeling step allows the network to correct itself, as it gradually becomes more certain, adding more samples to the mix. We refer to Section~\ref{subsec: experiments_ablation} for a concrete experiment. 

Algorithm~\ref{alg: algorithm} summarizes all the steps of the proposed method. We further refer to it as SCAN, i.e. Semantic Clustering by Adopting Nearest neighbors.

\begin{algorithm}[t]
\small{
\caption{Semantic Clustering by Adopting Nearest neighbors (SCAN)}
\label{alg: algorithm}
\begin{algorithmic}[1]
\State \textbf{Input:} Dataset $\mathcal{D}$, Clusters $\mathcal{C}$, Task $\tau$, Neural Nets $\Phi_\theta$ and $\Phi_\eta$, Neighbors $\mathcal{N}_\mathcal{D}=\{\}$.

\State Optimize $\Phi_\theta$ with task $\tau$. \Comment{Pretext Task Step, Sec.~\ref{subsec: method_neighbors}}
\For{$X_i \in \mathcal{D}$}
\State $\mathcal{N}_\mathcal{D} \leftarrow \mathcal{N}_{\mathcal{D}} \cup \mathcal{N}_{X_i}$, with $\mathcal{N}_{X_i} = K$ neighboring samples of $\Phi_\theta(X_i)$.
\EndFor
\While{SCAN-loss decreases} \Comment{Clustering Step, Sec.~\ref{subsec: method_loss}}
\State Update $\Phi_\eta$ with SCAN-loss, i.e. $\Lambda(\Phi_\eta(\mathcal{D}), \mathcal{N}_{\mathcal{D}}, C)$ in Eq.~\ref{eq:loss_objective}
\EndWhile 

\While{$Len(Y)$ increases} \Comment{Self-Labeling Step, Sec.~\ref{subsec: method_selflabeling}}
\State Y $\leftarrow$ ($\Phi_\eta(\mathcal{D}) >$ threshold)   
\State Update $\Phi_\eta$ with cross-entropy loss, i.e. $H(\Phi_\eta(\mathcal{D}), Y)$
\EndWhile

\State \textbf{Return:} $\Phi_\eta(\mathcal{D})$ \Comment{$\mathcal{D}$ is divided over $C$ clusters}
\end{algorithmic}
}
\end{algorithm}

\section{Experiments}
\label{sec:experiments}

\subsection{Experimental setup}
\label{subsec: experiments_datasets}
\noindent\textbf{Datasets.}
The experimental evaluation is performed on CIFAR10~\cite{CIFAR}, CIFAR100-20~\cite{CIFAR}, STL10~\cite{STL} and ImageNet~\cite{ImageNet}. We focus on the smaller datasets first. The results on ImageNet are discussed separately in Section~\ref{subsec: experiments_imagenet}. Some prior works~\cite{IIC,DAC,DEC,yang2016joint} trained and evaluated on the complete datasets. Differently, we train and evaluate using the train and val split respectively. Doing so, allows to study the generalization properties of the method for novel unseen examples. Note that this does not result in any unfair advantages compared to prior work. The results are reported as the mean and standard deviation from 10 different runs. Finally, all experiments are performed using the same backbone, augmentations, pretext task and hyperparameters.

\noindent\textbf{Training setup.}
We use a standard ResNet-18 backbone. For every sample, the 20 nearest neighbors are determined through an instance discrimination task based on noise contrastive estimation (NCE)~\cite{wu2018unsupervised}. We adopt the SimCLR~\cite{chen2020simple} implementation for the instance discrimination task on the smaller datasets, and the implementation from MoCo~\cite{chen2020improved} on ImageNet. The selected pretext task satisfies the feature invariance constraint from Equation~\ref{eq:invariance_objective} w.r.t. the transformations applied to augment the input images. In particular, every image is disentangled as a unique instance independent of the applied transformation. To speed up training, we transfer the weights, obtained from the pretext task to initiate the clustering step (Section~\ref{subsec: method_loss}). We perform the clustering step for 100 epochs using batches of size 128. The weight on the entropy term is set to $\lambda=5$. A higher weight avoids the premature grouping of samples early on during training. The results seem to be insensitive to small changes of $\lambda$. After the clustering step, we train for another 200 epochs using the self-labeling procedure with threshold $0.99$ (Section~\ref{subsec: method_selflabeling}). A weighted cross-entropy loss compensates for the imbalance between confident samples across clusters. The class weights are inversely proportional to the number of occurrences in the batch after thresholding. The network weights are updated through Adam~\cite{kingma2014adam} with learning rate $10^{-4}$ and weight decay $10^{-4}$. The images are strongly augmented by composing four randomly selected transformations from RandAugment~\cite{cubuk2020randaugment} during both the clustering and self-labeling steps. The transformation parameters are uniformly sampled between fixed intervals. For more details visit the supplementary materials. 

\noindent\textbf{Validation criterion}
During the clustering step, we select the best model based on the lowest loss. During the self-labeling step, we save the weights of the model when the amount of confident samples plateaus. We follow these practices as we do not have access to a labeled validation set. 

\subsection{Ablation studies}
\label{subsec: experiments_ablation}

\subsubsection{Method.}

\begin{figure}[t]
\begin{minipage}[t]{.31\linewidth} 
\centering
\includegraphics[width=\textwidth]{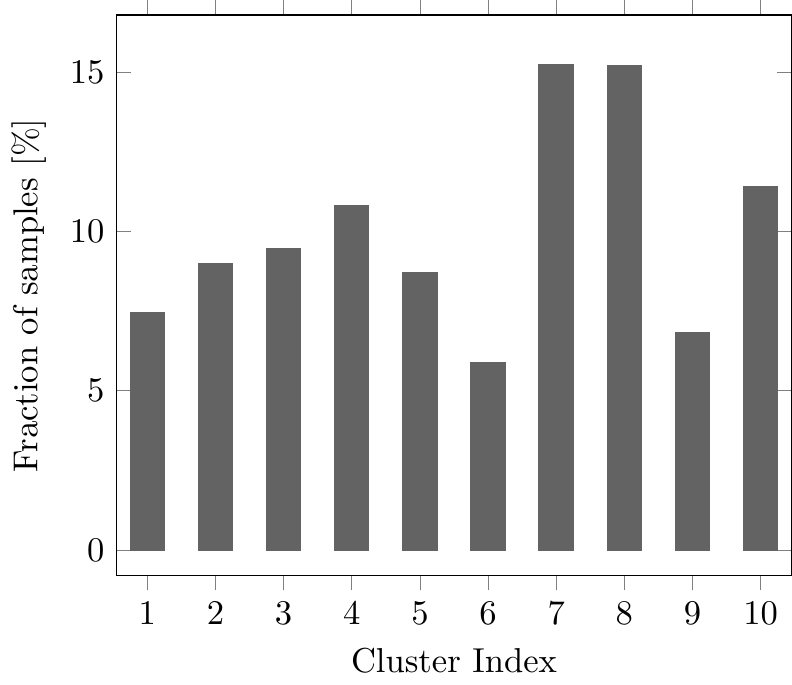}
\caption{K-means cluster assignments are imbalanced.}
\label{fig: kmeans}
\end{minipage} %
\hspace{0.02\linewidth}
\begin{minipage}[t]{.31\linewidth} 
\centering
\includegraphics[width=\textwidth]{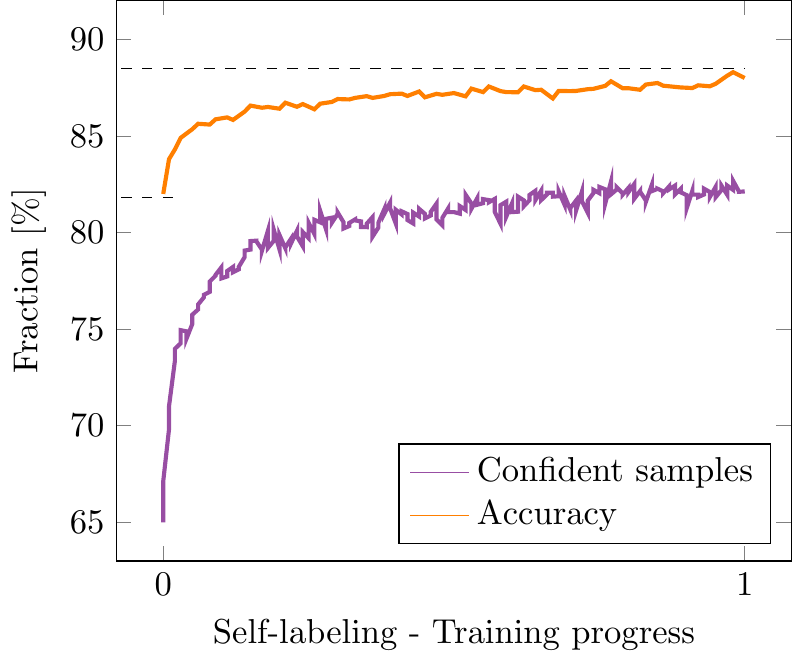}
\caption{Acc. and the number of confident samples during self-labeling.}
\label{fig: self_labeling}
\end{minipage} %
\hspace{0.02\linewidth}
\begin{minipage}[t]{.31\linewidth} 
\centering
\includegraphics[width=\textwidth]{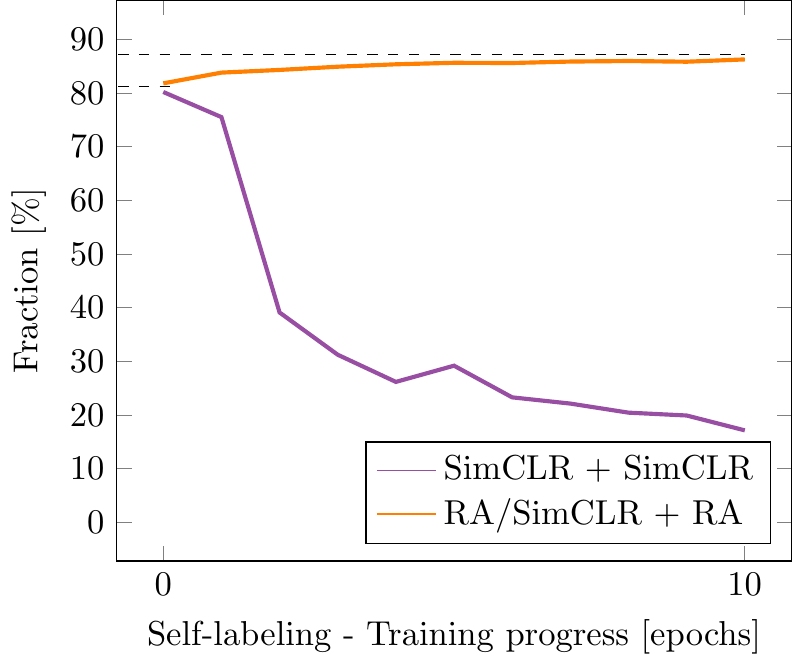}
\caption{Self-labeling with SimCLR or RandAugment augmentations.}
\label{fig: shift_augs}
\end{minipage} %
\end{figure}

We quantify the performance gains w.r.t. the different parts of our method through an ablation study on CIFAR10 in Table~\ref{table:ablation_method}. K-means clustering of the NCE pretext features results in the lowest accuracy ($65.9\%$), and is characterized by a large variance ($5.7\%$). This is to be expected since the cluster assignments can be imbalanced (Figure~\ref{fig: kmeans}), and are not guaranteed to align with the ground-truth classes. Interestingly, applying K-means to the pretext features outperforms prior state-of-the-art methods for unsupervised classification based on end-to-end learning schemes (see Sec.~\ref{subsec: experiments_sota}). This observation supports our primary claim, i.e. it is beneficial to separate feature learning from clustering. Updating the network weights through the SCAN-loss - while augmenting the input images through SimCLR transformations - outperforms K-means (+$15.9\%$). Note that the SCAN-loss is somewhat related to K-means, since both methods employ the pretext features as their prior to cluster the images. Differently, our loss avoids the cluster degeneracy issue. We also research the effect of using different augmentation strategies during training. Applying transformations from RandAgument (RA) to both the samples and their mined neighbors further improves the performance ($78.7\%$ vs. $81.8\%$). We hypothesize that strong augmentations help to reduce the solution space by imposing additional invariances.

\begin{figure}[t]
\begin{minipage}[t]{.31\linewidth} 
\centering
\includegraphics[width=\textwidth]{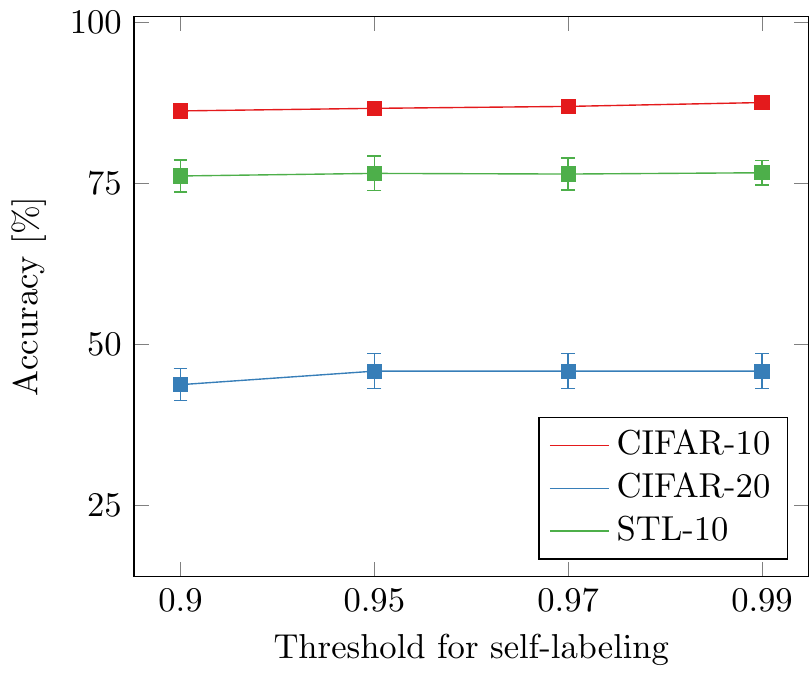}
\caption{Ablation threshold during self-labeling step.}
\label{fig: ablation_threshold}
\end{minipage} %
\hspace{0.02\linewidth}
\begin{minipage}[t]{.31\linewidth} 
\centering
\includegraphics[width=\textwidth]{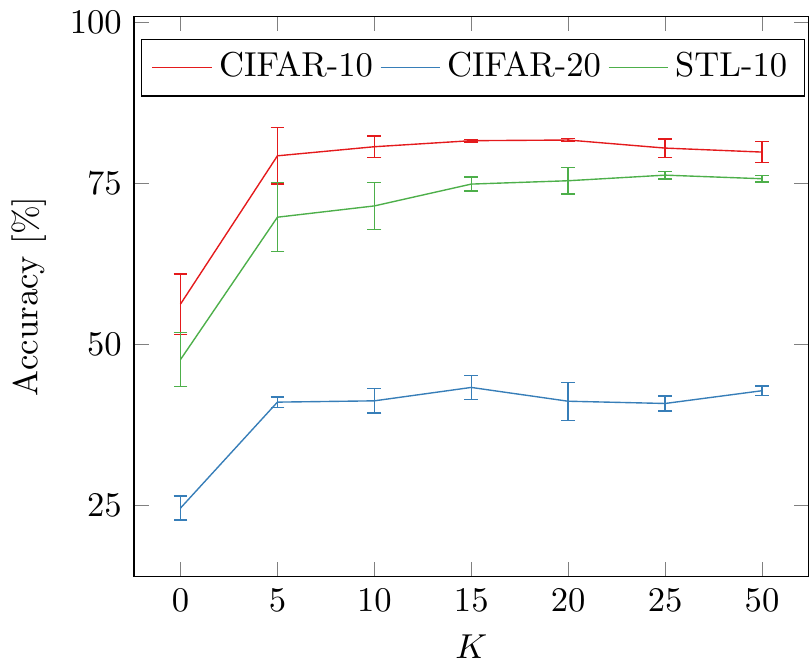}
\caption{Influence of the used number of neighbors $K$.}
\label{fig: topk}
\end{minipage} %
\hspace{0.02\linewidth}
\begin{minipage}[t]{.31\linewidth} 
\centering
\includegraphics[width=\textwidth]{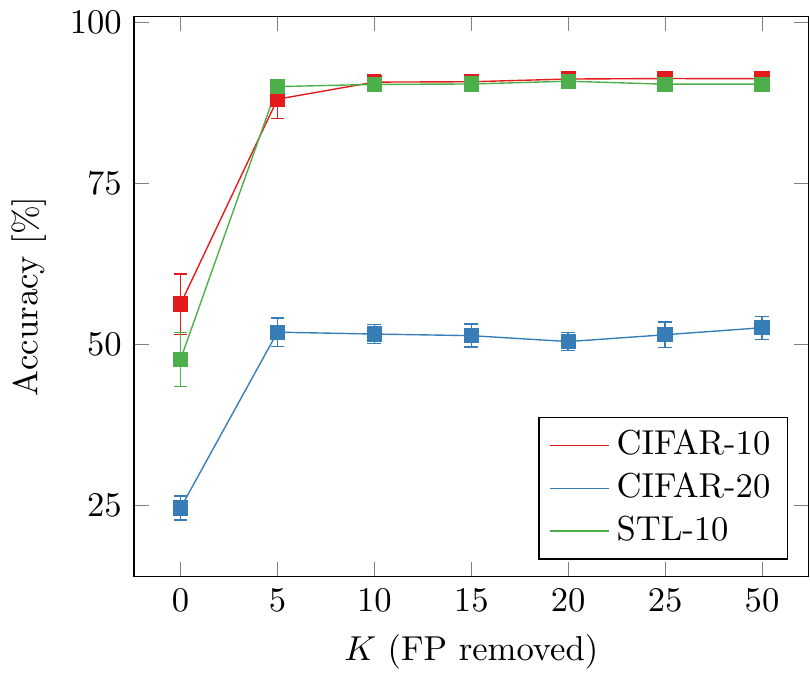}
\caption{Results without false positives in the nearest neighbors.}
\label{fig: convergence}
\end{minipage} %
\end{figure}

Fine-tuning the network through self-labeling further enhances the quality of the cluster assignments ($81.8\%$ to $87.6\%$). During self-labeling, the network corrects itself as it gradually becomes more confident~(see Figure~\ref{fig: self_labeling}). Importantly, in order for self-labeling to be successfully applied, a shift in augmentations is required (see Table~\ref{table:ablation_method} or Figure~\ref{fig: shift_augs}). We hypothesize that this is required to prevent the network from overfitting on already well-classified examples. Finally, Figure~\ref{fig: ablation_threshold} shows that self-labeling procedure is not sensitive to the threshold's value.

\setlength{\tabcolsep}{4pt}
\begin{table}[t]
\begin{minipage}[t]{0.5\linewidth}
\scriptsize
\begin{center}
\caption{Ablation Method CIFAR10}
\label{table:ablation_method}
\begin{tabular}{lcc}
\toprule
\textbf{Setup} & \ \textbf{ACC} \\
 &  (Avg $\pm$ Std) \\
\midrule
Pretext + K-means & $65.9\pm5.7$ \\
SCAN-Loss (SimCLR) & $78.7\pm1.7$ \\
\hspace{0.1in} (1) Self-Labeling (SimCLR) & $10.0\pm0$ \\
\hspace{0.1in} (2) Self-Labeling (RA) & $87.4\pm1.6$ \\
SCAN-Loss (RA) & $81.8\pm1.7$ \\
\hspace{0.1in} (1) Self-Labeling (RA) & $87.6\pm0.4$ \\
\bottomrule
\end{tabular}
\end{center}
\end{minipage}
\hspace*{\fill}
\begin{minipage}[t]{0.5\linewidth}
\scriptsize
\begin{center}
\caption{Ablation Pretext CIFAR10}
\label{table: ablation_pretext}
\begin{tabular}{lcc}
\toprule
\textbf{Pretext Task} & \textbf{Clustering}& \textbf{ACC}\\
  & & (Avg $\pm$ Std)  \\
\midrule
RotNet~\cite{RotNet}  & K-means & $27.1\pm2.1$ \\
  & SCAN & $74.3\pm3.9$ \\
Inst. discr.~\cite{wu2018unsupervised} & K-means & $52.0\pm4.6$\\
 & SCAN &  $83.5\pm4.1$\\
Inst. discr.~\cite{chen2020simple} & K-means & $65.9\pm5.7$ \\
 & SCAN &  $87.6\pm0.4$\\
\bottomrule
\end{tabular}
\end{center}
\end{minipage}
\end{table}
\setlength{\tabcolsep}{1.4pt}


\subsubsection{Pretext task.}
We study the effect of using different pretext tasks to mine the nearest neighbors. In particular we consider two different implementations of the instance discrimination task from before~\cite{wu2018unsupervised,chen2020simple}, and RotNet~\cite{RotNet}. The latter trains the network to predict image rotations. As a consequence, the distance between an image $X_i$ and its augmentations $T[X_i]$ is not minimized in the embedding space of a model pretrained through RotNet (see Equation~\ref{eq:invariance_objective}). Differently, the instance discrimintation task satisfies the invariance criterion w.r.t. the used augmentations. Table~\ref{table: ablation_pretext} shows the results on CIFAR10. 

First, we observe that the proposed method is not tied to a specific pretext task. All cases report high accuracy ($>70\%$). Second, pretext tasks that satisfy the invariance criterion are better suited to mine the nearest neighbors, i.e. $83.5\%$ and $87.6\%$ for inst. discr. versus $74.3\%$ for RotNet. This confirms our hypothesis from Section~\ref{subsec: method_neighbors}, i.e. it is beneficial to choose a pretext task which imposes invariance between an image and its augmentations. 

\subsubsection{Number of neighbors.}
Figure~\ref{fig: topk} shows the influence of using a different number of nearest neighbors $K$ during the clustering step. The results are not very sensitive to the value of $K$, and even remain stable when increasing $K$ to 50. This is beneficial, since we do not have to fine-tune the value of $K$ on very new dataset. In fact, both robustness and accuracy improve when increasing the value of $K$ upto a certain value. We also consider the corner case $K=0$, when only enforcing consistent predictions for images and their augmentations. the performance decreases on all three datasets compared to $K=5$, $56.3\%$ vs $79.3\%$ on CIFAR10, $24.6\%$ vs $41.1\%$ on CIFAR100-20 and $47.70 \%$ vs $69.8\%$ on STL10. This confirms that better representations can be learned by also enforcing coherent predictions between a sample and its nearest neighbors. 

\subsubsection{Convergence.}
Figure~\ref{fig: convergence} shows the results when removing the false positives from the nearest neighbors, i.e. sample-pairs which belong to a different class. The results can be considered as an upper-bound for the proposed method in terms of classification accuracy. A desirable characteristic is that the clusters quickly align with the ground truth, obtaining near fully-supervised performance on CIFAR10 and STL10 with a relatively small increase in the number of used neighbors $K$. The lower performance improvement on CIFAR100-20 can be explained by the ambiguity of the superclasses used to measure the accuracy. For example, there is not exactly one way to group categories like omnivores or carnivores together. 

\subsection{Comparison with the state-of-the-art}
\label{subsec: experiments_sota}
\noindent\textbf{Comparison.} Table~\ref{table:sota} compares our method to the state-of-the-art on three different benchmarks. We evaluate the results based on clustering accuracy (ACC), normalized mutual information (NMI) and adjusted rand index (ARI). The proposed method consistently outperforms prior work by large margins on all three metrics, e.g. $+26.6\%$ on CIFAR10, $+25.0\%$ on CIFAR100-20 and $+21.3\%$ on STL10 in terms of accuracy. We also compare with the state-of-the-art in representation learning~\cite{chen2020simple} (Pretext + K-means). As shown in Section~\ref{subsec: experiments_ablation}, our method outperforms the application of K-means on the pretext features. Finally, we also include results when tackling the problem in a fully-supervised manner. Our model obtains close to supervised performance on CIFAR-10 and STL-10. The performance gap is larger on CIFAR100-20, due to the use of superclasses. 

\noindent\textbf{Other advantages.} In contrast to prior work~\cite{DAC,IIC,hu2017learning}, we did not have to perform any dataset specific fine-tuning. Furthermore, the results on CIFAR10 can be obtained within 6 hours on a single GPU. As a comparison, training the model from~\cite{IIC} requires at least a day of training time.

\setlength{\tabcolsep}{4pt}
\begin{table}[t]
\begin{center}
\caption{State-of-the-art comparison: We report the averaged results for 10 different runs after the clustering ($*$) and self-labeling steps ($\dagger$), and the best model. Opposed to prior work, we train and evaluate using the train and val split respectively, instead of using the full dataset for both training and testing.}
\label{table:sota}
\resizebox{\columnwidth}{!}{
\begin{tabular}{@{}l c ccc c ccc c ccc@{}}
\toprule
\textbf{Dataset} && \multicolumn{3}{c}{\textbf{CIFAR10}} && \multicolumn{3}{c}{\textbf{CIFAR100-20}} && \multicolumn{3}{c}{\textbf{STL10}} \\
\cmidrule{3-5} \cmidrule{7-9} \cmidrule{11-13}
\textbf{Metric} && ACC & NMI & ARI && ACC & NMI & ARI&& ACC & NMI & ARI \\
\midrule
\noalign{\smallskip}

K-means~\cite{wang2015optimized}     && 22.9 & 8.7   & 4.9  && 13.0  & 8.4  & 2.8 && 19.2 & 12.5 & 6.1  \\
SC~\cite{zelnik2005self}             && 24.7 & 10.3  & 8.5  && 13.6  & 9.0  & 2.2 && 15.9 & 9.8  & 4.8  \\
Triplets~\cite{schultz2004learning}  && 20.5 & --    & --   && 9.94  & --   & --  && 24.4 & --   & --   \\
JULE~\cite{yang2016joint}            && 27.2 & 19.2  & 13.8 && 13.7  & 10.3 & 3.3 && 27.7 & 18.2 & 16.4 \\
AEVB~\cite{kingma2013auto}           && 29.1 & 24.5  & 16.8 && 15.2  & 10.8 & 4.0 && 28.2 & 20.0 & 14.6 \\
SAE~\cite{ng2011sparse}              && 29.7 & 24.7  & 15.6 && 15.7  & 10.9 & 4.4 && 32.0 & 25.2 & 16.1 \\
DAE~\cite{vincent2010stacked}        && 29.7 & 25.1  & 16.3 && 15.1  & 11.1 & 4.6 && 30.2 & 22.4 & 15.2 \\
SWWAE~\cite{zhao2015stacked}         && 28.4 & 23.3  & 16.4 && 14.7  & 10.3 & 3.9 && 27.0 & 19.6 & 13.6 \\
AE~\cite{bengio2007greedy}           && 31.4 & 23.4  & 16.9 && 16.5  & 10.0 & 4.7 && 30.3 & 25.0 & 16.1 \\
GAN~\cite{radford2015unsupervised}   && 31.5 & 26.5  & 17.6 && 15.1  & 12.0 & 4.5 && 29.8 & 21.0 & 13.9 \\
DEC~\cite{DEC}                       && 30.1 & 25.7  & 16.1 && 18.5  & 13.6 & 5.0 && 35.9 & 27.6 & 18.6 \\
ADC~\cite{haeusser2018associative}   && 32.5 & --    & --   && 16.0  & --   & --  && 53.0 & --   & --   \\
DeepCluster~\cite{DeepCluster}       && 37.4 & --    & --   && 18.9  & --   & --  && 33.4 & --   & --   \\
DAC~\cite{DAC}                       && 52.2 & 40.0 & 30.1 && 23.8& 18.5 & 8.8 && 47.0 & 36.6 & 25.6 \\
IIC~\cite{IIC} && \underline{61.7} & \underline{51.1}  & \underline{41.1} && \underline{25.7}  & \underline{22.5} & \underline{11.7}&& \underline{59.6} & \underline{49.6} & \underline{39.7} \\
\midrule
Supervised  && 93.8 &  86.2 & 87.0   && 80.0 & 68.0 & 63.2 && 80.6 & 65.9 & 63.1\\
Pretext~\cite{chen2020simple} + K-means && $65.9\pm5.7$ & $59.8\pm2.0$ & $50.9\pm3.7$   && $39.5\pm1.9$ & $40.2\pm1.1$ & $23.9\pm1.1$ && $65.8\pm5.1$ & $60.4\pm2.5$ & $50.6\pm4.1$\\
\textbf{SCAN$^*$ (Avg $\pm$ Std) }  &&$81.8\pm0.3$ & $71.2\pm 0.4$ & $66.5\pm0.4 $&& $42.2\pm3.0$ & $44.1\pm1.0$ & $26.7\pm1.3$ && $75.5\pm2.0$ & $65.4\pm1.2$&$59.0\pm1.6$  \\
\textbf{SCAN$^\dagger$ (Avg $\pm$ Std) } &&$87.6\pm0.4$ & $78.7\pm 0.5$ & $75.8\pm0.7 $&& $45.9\pm2.7$ & $46.8\pm1.3$ & $30.1\pm2.1$ && $76.7\pm1.9$ & $68.0\pm1.2$&$61.6\pm1.8$  \\
\textbf{SCAN$^\dagger$ (Best)} && \textbf{88.3} &\textbf{79.}7 & \textbf{77.2} && \textbf{50.7} & \textbf{48.6} & \textbf{33.3} && \textbf{80.9} & \textbf{69.8} & \textbf{64.6} \\
\midrule
\textbf{SCAN$^\dagger$ (Overcluster)} &&$86.2\pm0.8$ & $77.1\pm 0.1$ & $73.8\pm1.4 $&& $55.1\pm1.6$ & $50.0\pm1.1$ & $35.7\pm1.7$ && $76.8\pm1.1$ & $65.6\pm0.8$&$58.6\pm1.6$  \\
\bottomrule
\end{tabular}
}
\end{center}
\end{table}
\setlength{\tabcolsep}{1.4pt}

\subsection{Overclustering}
\label{subsec: experiments_overclustering}
So far we assumed to have knowledge about the number of ground-truth classes. The method predictions were evaluated using a hungarian matching algorithm. However, what happens if the number of clusters does not match the number of ground-truth classes anymore. Table~\ref{table:sota} reports the results when we overestimate the number of ground-truth classes by a factor of 2, e.g. we cluster CIFAR10 into 20 rather than 10 classes. The classification accuracy remains stable for CIFAR10 ($87.6\%$ to $86.2\%$) and STL10 ($76.7\%$ to $76.8\%$), and improves for CIFAR100-20 ($45.9\%$ to $55.1\%$)\footnote{Since the overclustering case is evaluated using a many-to-one mapping, a direct comparison is not entirely fair. Still, we provide the comparison as an indication.}. We conclude that the approach does not require knowledge of the exact number of clusters. We hypothesize that the increased performance on CIFAR100-20 is related to the higher intra-class variance. More specifically, CIFAR100-20 groups multiple object categories together in superclasses. In this case, an overclustering is better suited to explain the intra-class variance. 

\subsection{ImageNet}
\label{subsec: experiments_imagenet}
\subsubsection{Setup.}
We consider the problem of unsupervised image classification on the large-scale ImageNet dataset~\cite{ImageNet}. We first consider smaller subsets of 50, 100 and 200 randomly selected classes. The sets of 50 and 100 classes are subsets of the 100 and 200 classes respectively. Additional details of the training setup can be found in the supplementary materials. 

\subsubsection{Quantitative evaluation.}
Table~\ref{table: imagenet_subsets} compares our results against applying K-means on the pretext features from MoCo~\cite{chen2020improved}. Surprisingly, the application of K-means already performs well on this challenging task. We conclude that the pretext features are well-suited for the down-stream task of semantic clustering. Training the model with the SCAN-loss again outperforms the application of K-means. Also, the results are further improved when fine-tuning the model through self-labeling. We do not include numbers for the prior state-of-the-art~\cite{IIC}, since we could not obtain convincing results on ImageNet when running the publicly available code. We refer the reader to the supplementary materials for additional qualitative results on ImageNet-50.

\setlength{\tabcolsep}{4pt}
\begin{table}[t]
\begin{center}
\caption{Validation set results for 50, 100 and 200 randomly selected classes from ImageNet. The results with K-means were obtained using the pretext features from MoCo~\cite{chen2020improved}. We provide the results obtained by our method after the clustering step $(*)$, and after the self-labeling step $(\dagger)$.}
\label{table: imagenet_subsets}
\resizebox{\columnwidth}{!}{
\begin{tabular}{@{}l c cccc c cccc c cccc@{}}
\toprule
\textbf{ImageNet} && \multicolumn{4}{c}{\textbf{50 Classes}} && \multicolumn{4}{c}{\textbf{100 Classes}} && \multicolumn{4}{c}{\textbf{200 Classes}} \\
\cmidrule{3-6} \cmidrule{8-11} \cmidrule{13-16}
\textbf{Metric} && Top-1 & Top-5 & NMI & ARI && Top-1 & Top-5 & NMI & ARI && Top-1 & Top-5 & NMI & ARI \\
\midrule
\noalign{\smallskip}
\textbf{K-means} && 65.9& - & 77.5 & 57.9 && 59.7 & - & 76.1 & 50.8 && 52.5& - & 75.5& 43.2 \\
\textbf{SCAN$^{*}$} && 75.1 & 91.9 & 80.5 & 63.5 && 66.2 & 88.1 & 78.7 & 54.4 && 56.3 & 80.3 & 75.7 & 44.1 \\
\textbf{SCAN$^{\dagger}$}  && 76.8 & 91.4 & 82.2 & 66.1 && 68.9 & 86.1 & 80.8 & 57.6 && 58.1 & 80.6 & 77.2 & 47.0 \\
\bottomrule
\end{tabular}
}
\end{center}
\end{table}
\setlength{\tabcolsep}{1.4pt}

\begin{figure}[t]
\begin{minipage}[t]{0.48\linewidth}
\centering
\includegraphics[width=1.00\textwidth]{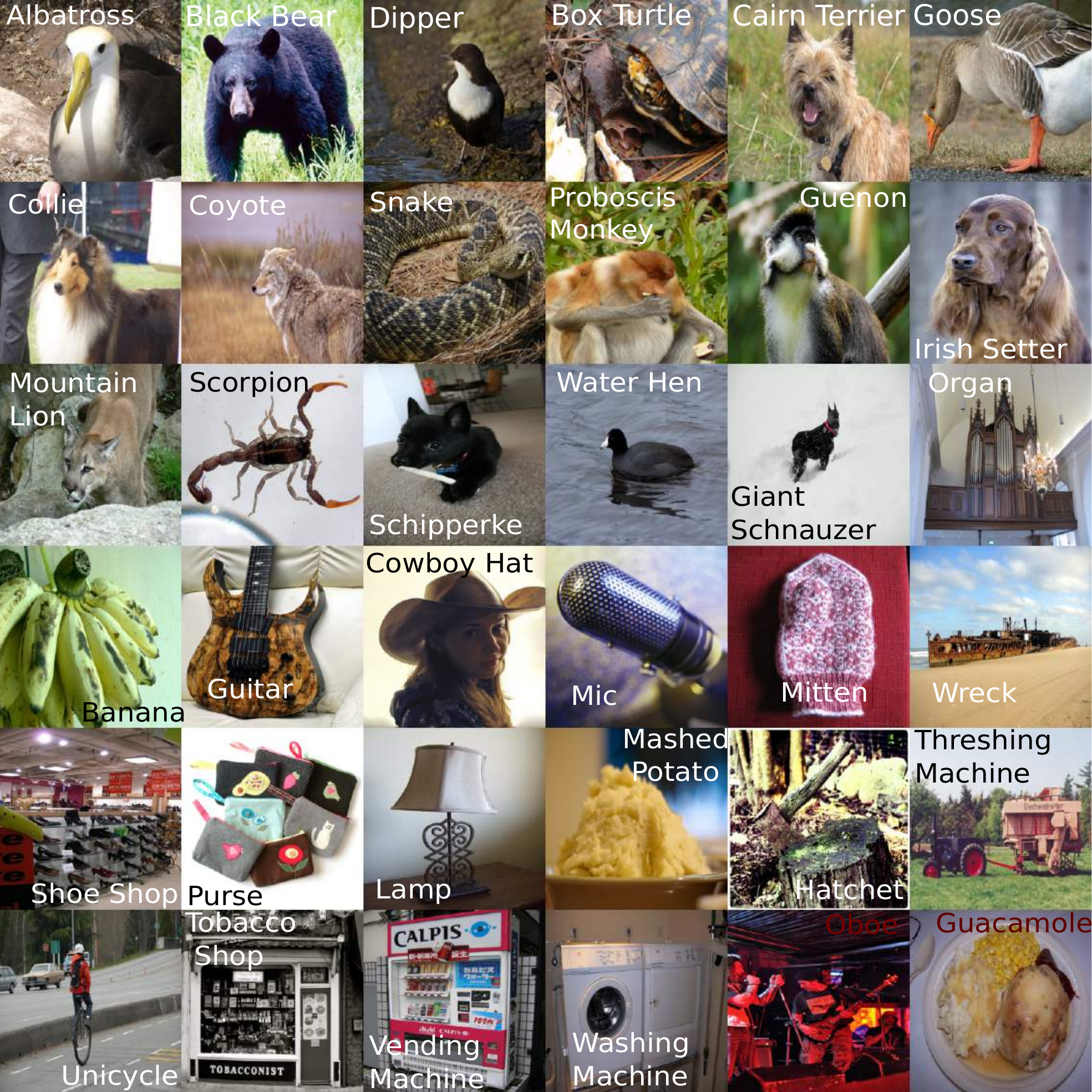}
\caption{Prototypes obtained by sampling the confident samples.}
\label{fig: prototypes}
\end{minipage}
\hspace*{\fill}
\begin{minipage}[t]{0.48\linewidth}
\centering
\includegraphics[width=1.00\textwidth]{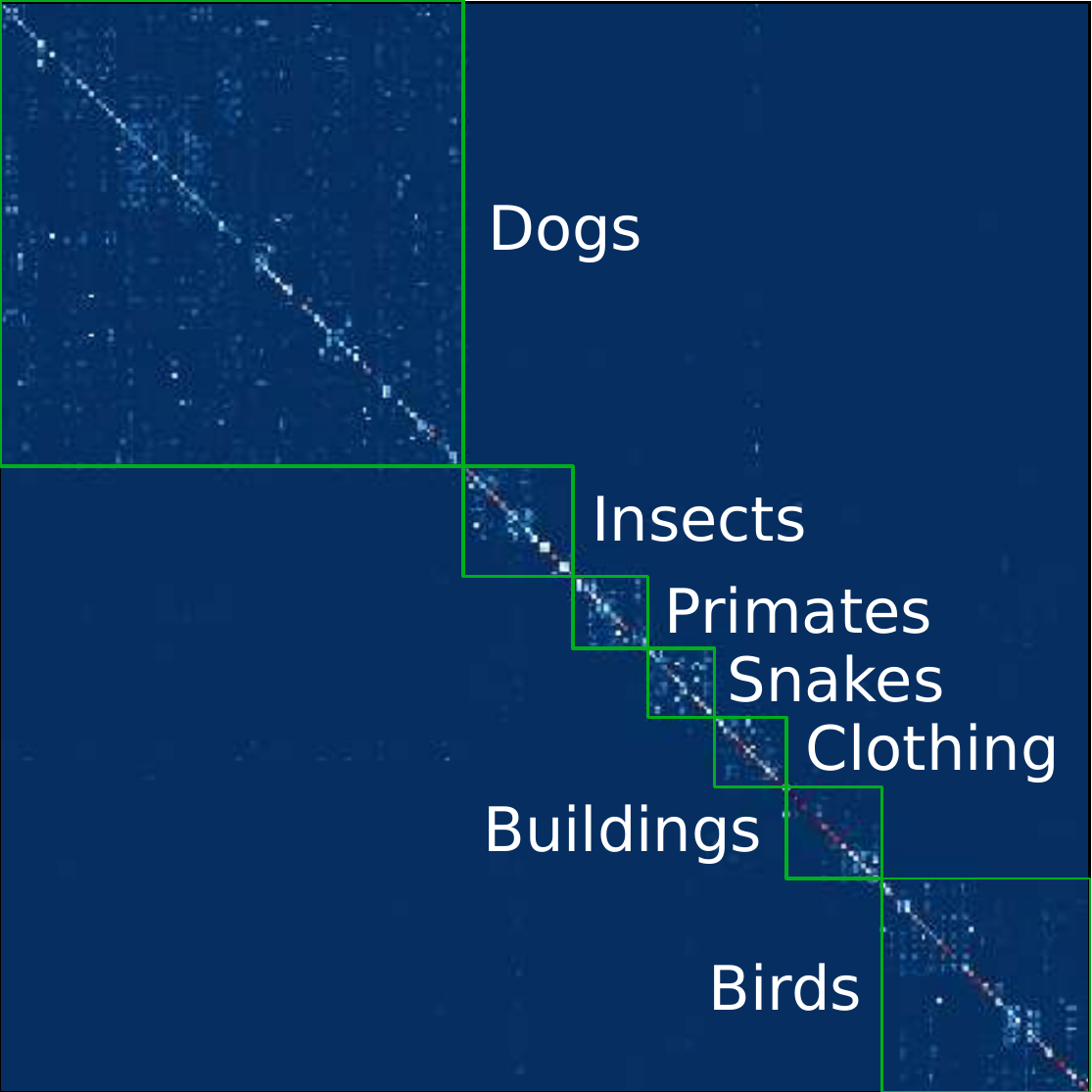}
\caption{Zoom on seven superclasses in the confusion matrix on ImageNet.}
\label{fig: imagenet_confusion}
\end{minipage}
\end{figure}

\subsubsection{Prototypical behavior.}
We visualize the different clusters after training the model with the SCAN-loss. Specifically, we find the samples closest to the mean embedding of the top-10 most confident samples in every cluster. The results are shown together with the name of the matched ground-truth classes in Fig.~\ref{fig: prototypes}. Importantly, we observe that the found samples align well with the classes of the dataset, except for 'oboe' and 'guacamole' (red). Furthermore, the discriminative features of each object class are clearly present in the images. Therefore, we regard the obtained samples as "prototypes" of the various clusters. Notice that the performed experiment aligns well with prototypical networks~\cite{snell2017prototypical}.

\begin{figure}
\centering
\includegraphics[width=\textwidth]{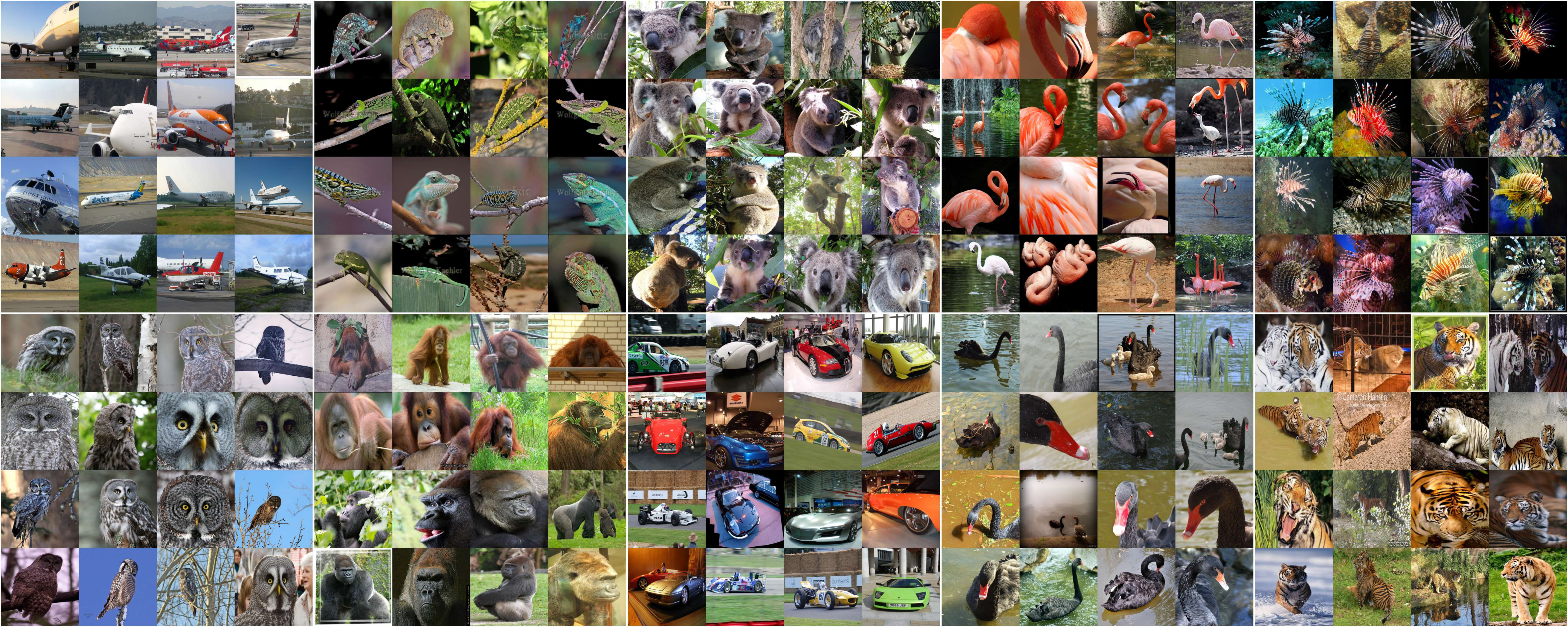}
\caption{Clusters extracted by our model on ImageNet (more in supplementary).}
\label{fig: imagenet_clusters}
\end{figure}

\subsubsection{ImageNet - 1000 classes.}
Finally, the model is trained on the complete ImageNet dataset. Figure~\ref{fig: imagenet_clusters} shows images from the validation set which were assigned to the same cluster by our model. The obtained clusters are semantically meaningful, e.g. planes, cars and primates. Furthermore, the clusters capture a large variety of different backgrounds, viewpoints, etc. We conclude that (to a large extent) the model predictions are invariant to image features which do not alter the semantics. On the other hand, based on the ImageNet ground-truth annotations, not all sample pairs should have been assigned to the same cluster. For example, the ground-truth annotations discriminate between different primates, e.g. chimpanzee, baboon, langur, etc. We argue that there is not a single correct way of categorizing the images according to their semantics in case of ImageNet. Even for a human annotator, it is not straightforward to cluster each image according to the ImageNet classes without prior knowledge. 

Based on the ImageNet hierarchy we select class instances of the following superclasses: dogs, insects, primates, snake, clothing, buildings and birds. Figure~\ref{fig: imagenet_confusion} shows a confusion matrix of the selected classes. The confusion matrix has a block diagonal structure. The results show that the misclassified examples tend to be assigned to other clusters from within the same superclass, e.g. the model confuses two different dog breeds. We conclude that the model has learned to group images with similar semantics together, while its prediction errors can be attributed to the lack of annotations which could disentangle the fine-grained differences between some classes. 

Finally, Table~\ref{table: imagenet} compares our method against recent semi-supervised learning approaches when using $1\%$ of the images as labelled data. We obtain the following quantitative results on ImageNet: Top-1: $39.9\%$, Top-5: $60.0\%$, NMI: $72.0\%$, ARI: $27.5\%$. Our method outperforms several semi-supervised learning approaches, without using labels. This further demonstrates the strength of our approach. 

\setlength{\tabcolsep}{4pt}
\begin{table}
\begin{center}
\caption{Comparison with supervised, and semi-supervised learning methods using $1\%$ of the labelled data on ImageNet.}
\label{table: imagenet}
\scriptsize
\begin{tabular}{@{}l l ccc@{}}
\toprule
\textbf{Method} & \textbf{Backbone} & \textbf{Labels} & \textbf{Top-1} & \textbf{Top-5} \\
\midrule
Supervised Baseline & ResNet-50 & $\checkmark$ & 25.4 & 48.4 \\
Pseudo-Label & ResNet-50 & $\checkmark$ & - & 51.6 \\
VAT + Entropy Min.~\cite{zhai2019s4l} & ResNet-50 & $\checkmark$ & - & 47.0 \\
InstDisc~\cite{wu2018unsupervised} & ResNet-50 & $\checkmark$ & - & 39.2 \\
BigBiGAN~\cite{donahue2019large} & ResNet-50(4x) & $\checkmark$ & - & 55.2 \\
PIRL~\cite{PIRL} & ResNet-50 & $\checkmark$ & - & 57.2 \\
CPC v2~\cite{henaff2019data} & ResNet-161 & $\checkmark$ & 52.7 & 77.9 \\
SimCLR~\cite{chen2020simple} & ResNet-50 & $\checkmark$ & 48.3 & 75.5 \\
\midrule
SCAN (Ours) & ResNet-50 & \xmark & 39.9 & 60.0 \\
\bottomrule
\end{tabular}
\end{center}
\end{table}
\setlength{\tabcolsep}{1.4pt}

\section{Conclusion}
\label{sec:conclusion}
We presented a novel framework to unsupervised image classification. The proposed approach comes with several advantages relative to recent works which adopted an end-to-end strategy. Experimental evaluation shows that the proposed method outperforms prior work by large margins, for a variety of datasets. Furthermore, positive results on ImageNet demonstrate that semantic clustering can be applied to large-scale datasets. Encouraged by these findings, we believe that our approach admits several extensions to other domains, e.g. semantic segmentation, semi-supervised learning and few-shot learning. 

\subsubsection{Acknowledgment.}
The authors thankfully acknowledge support by Toyota via the TRACE project and MACCHINA (KU Leuven, C14/18/065). Furthermore, we would like to thank Xu Ji for her valuable insights and comments. Finally, we thank Kevis-Kokitsi Maninis, Jonas Heylen and Mark De Wolf for their feedback.

\bibliographystyle{splncs04}
\bibliography{citations_suppl}

\newpage

\begin{center}
\Large{\textbf{Supplementary Material}}
\end{center}

\setcounter{section}{0}
\renewcommand\thesection{\Alph{section}}
\setcounter{figure}{0}
\setcounter{table}{0}
\renewcommand{\thefigure}{S\arabic{figure}}
\renewcommand{\thetable}{S\arabic{table}}

\section{Smaller datasets}
\label{sec:results}
We include additional qualitative results on the smaller datasets, i.e. CIFAR10~\cite{CIFAR}, CIFAR100-20~\cite{CIFAR} and STL10~\cite{STL}. We used the models from the state-of-the-art comparison. 

\subsection{Prototypical examples}
\begin{figure}
    \centering
    \caption{Prototype images on the smaller datasets.}
    \label{fig:prototypes}
    \begin{subfigure}{\textwidth} 
    \caption{CIFAR10}
    \includegraphics[width=\linewidth]{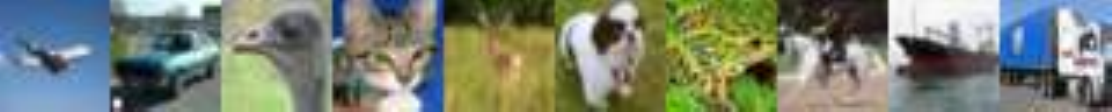}
    \end{subfigure}
    \newline
    \begin{subfigure}{\textwidth} 
    \caption{STL10}
    \includegraphics[width=\linewidth]{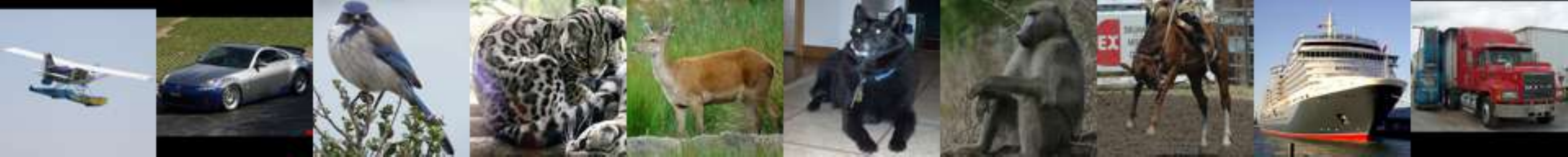}
    \end{subfigure}
    \begin{subfigure}{\textwidth} 
    \caption{CIFAR100-20}
    \includegraphics[width=\linewidth]{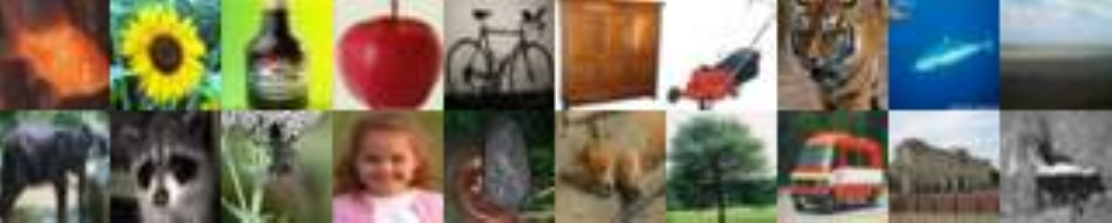}
    \end{subfigure}
\end{figure}

Figure~\ref{fig:prototypes} visualizes a prototype image for every cluster on CIFAR10, CIFAR100-20 and STL-10. The object of interest is clearly recognizable in the images. It is worth noting that the prototypical examples on CIFAR10 and STL10 can be matched with the ground-truth classes of the dataset. This is not the case for CIFAR100-20, e.g. \textit{bus} and \textit{bicycle} belong to the \textit{vehicles 1} ground-truth class. This behavior can be easily understood since CIFAR-20 makes use of superclasses. As a consequence, it is difficult to explain the intra-class variance from visual appearance alone. Interestingly, we can reduce this mismatch through overclustering (see Sec 3.4.). 

\begin{figure}
    \centering
    \caption{Low confidence predictions.}
    \label{fig: low_confidence}
    \begin{subfigure}{\textwidth} 
    \caption{CIFAR10}
    \includegraphics[width=\linewidth]{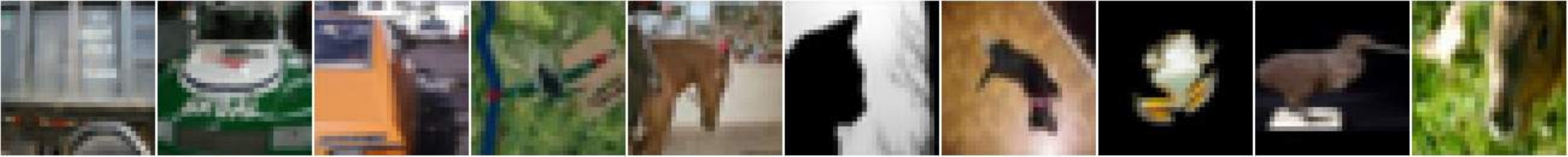}
    \end{subfigure}
    \newline
    \begin{subfigure}{\textwidth} 
    \caption{STL10}
    \includegraphics[width=\linewidth]{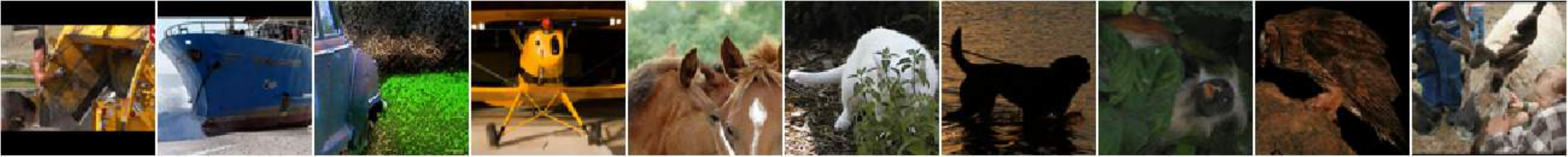}
    \end{subfigure}
    \begin{subfigure}{\textwidth} 
    \caption{CIFAR100-20}
    \includegraphics[width=\linewidth]{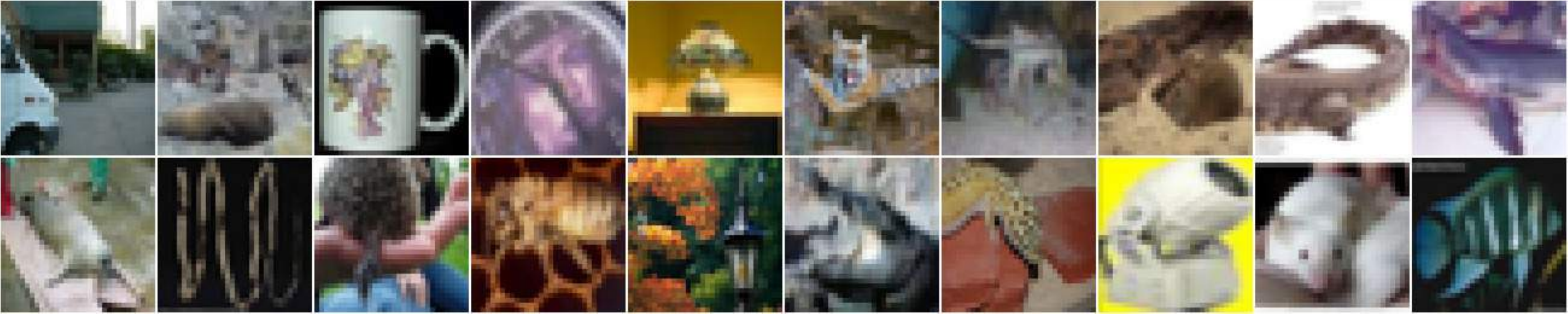}
    \end{subfigure}
\end{figure}

\subsection{Low confidence examples}
Figure~\ref{fig: low_confidence} shows examples for which the network produces low confidence predictions. In most cases, it is hard to determine the correct class label. The difficult examples include objects which are: only partially visible, occluded, under bad lighting conditions, etc.

\section{ImageNet}
\subsection{Training setup}
We summarize the training setup for ImageNet below.

\subsubsection{Pretext Task}
Similar to our setup on the smaller datasets, we select instance discrimination as our pretext task. In particular, we use the implementation from MoCo~\cite{chen2020improved}. We use a ResNet-50 model as backbone. 

\subsubsection{Clustering Step}
We freeze the backbone weights during the clustering step, and only train the final linear layer using the SCAN-loss. More specifically, we train ten separate linear heads in parallel. When initiating the self-labeling step, we select the head with the lowest loss to continue training. Every image is augmented using augmentations from SimCLR~\cite{chen2020simple}. We reuse the entropy weight from before (5.0), and train with batches of size 512, 1024 and 1024 on the subsets of 50, 100 and 200 classes respectively. We use an SGD optimizer with momentum $0.9$ and initial learning rate $5.0$. The model is trained for 100 epochs. On the full ImageNet dataset, we increase the batch size and learning rate to $4096$ and $30.0$ respectively, and decrease the number of neighbors to 20.

\subsubsection{Self-Labeling Step}
We use the strong augmentations from RandAugment to finetune the weights through self-labeling. The model weights are updated for 25 epochs using SGD with momentum $0.9$. The initial learning rate is set to $0.03$ and kept constant. Batches of size 512 are used. Importantly, the model weights are updated through an exponential moving average with $\alpha=0.999$. We did not find it necessary to apply class balancing in the cross-entropy loss. 

\subsection{ImageNet - Subsets}
\subsubsection{Confusion matrix}
\begin{figure}[ht]
\centering
\includegraphics[width=0.9\textwidth]{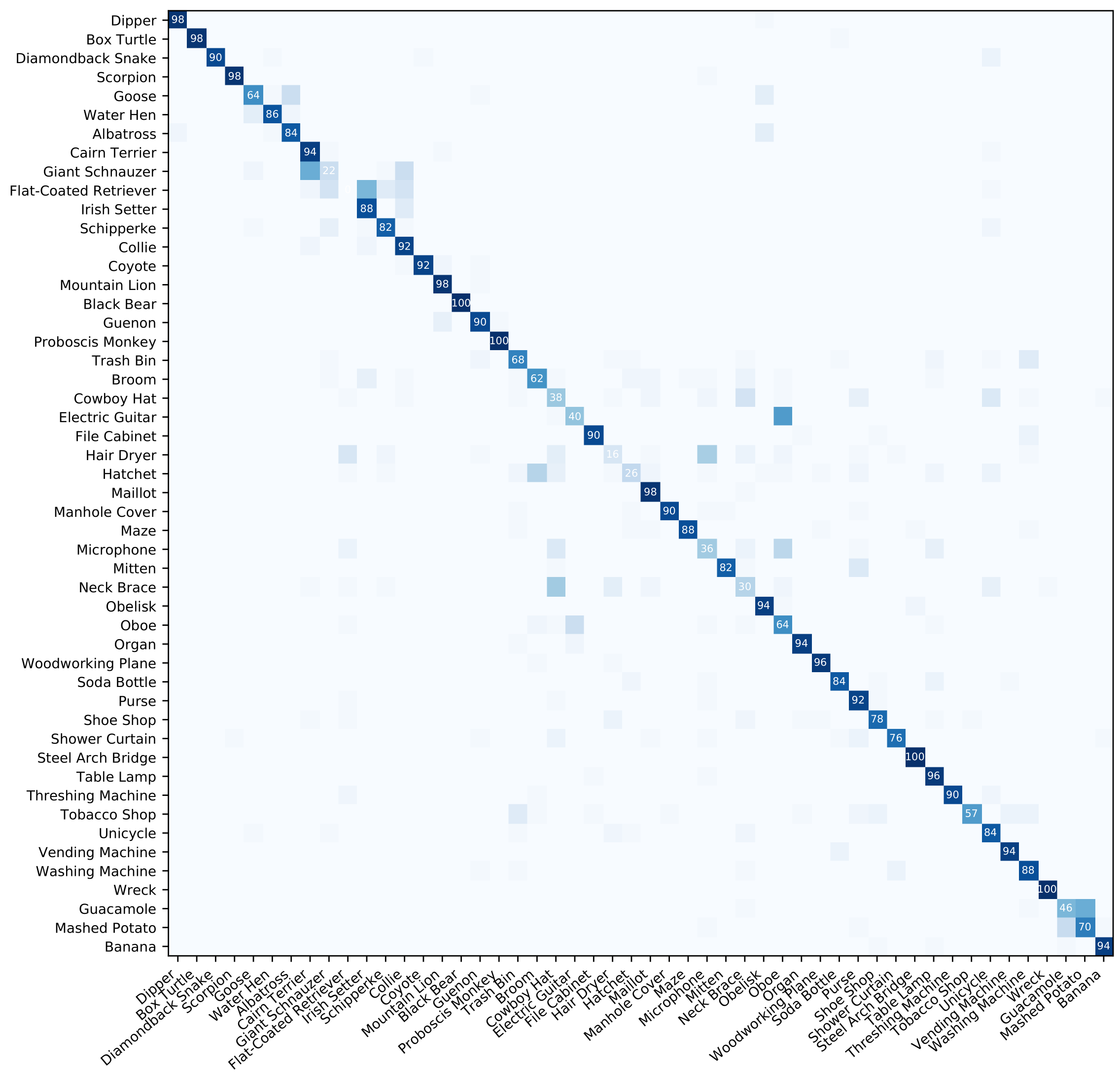}
\caption{Confusion matrix on ImageNet-50.}
\label{fig: confusion_imagenet}
\end{figure}

Figure~\ref{fig: confusion_imagenet} shows a confusion matrix on the ImageNet-50 dataset. Most of the mistakes can be found between classes that are hard to disentangle, e.g. \textit{'Giant Schnauzer'} and \textit{'Flat-coated Retriever'} are both black dog breeds, \textit{'Guacamole'} and \textit{'Mashed Potato'} are both food, etc. 
\subsubsection{Prototype examples}
\begin{figure}[t]
\centering
\includegraphics[width=\linewidth]{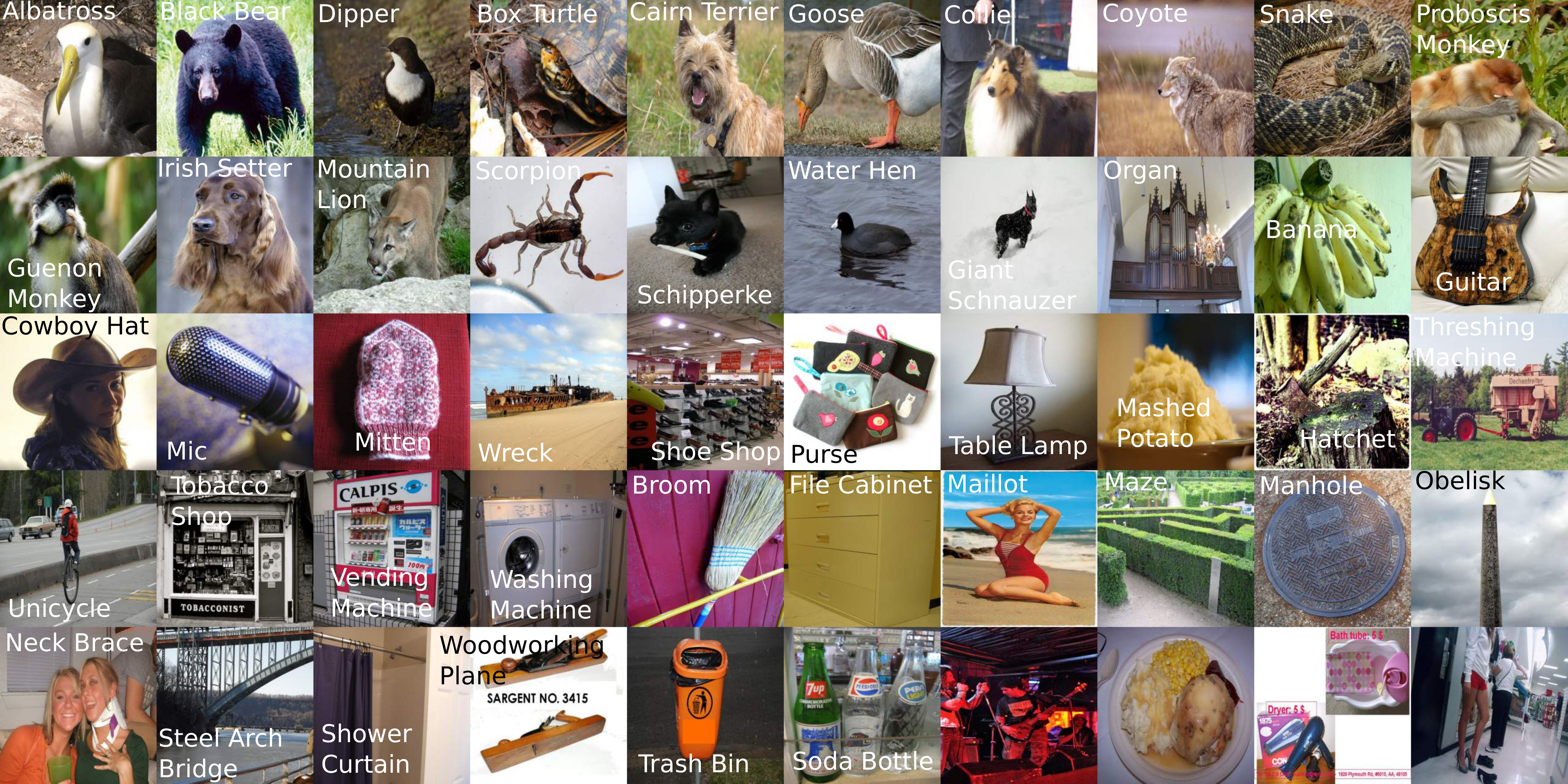}
\caption{Prototype images on ImageNet-50.}
\label{fig: prototypes_imagenet50}
\end{figure}

Figure~\ref{fig: prototypes_imagenet50} shows a prototype image for every cluster on the ImageNet-50 subset. This figure extends Figure 9 from the main paper. Remarkably, the vast majority of prototype images can be matched with one of the ground-truth classes. 

\subsubsection{Low confidence examples}
\begin{figure}
\includegraphics[width=\linewidth]{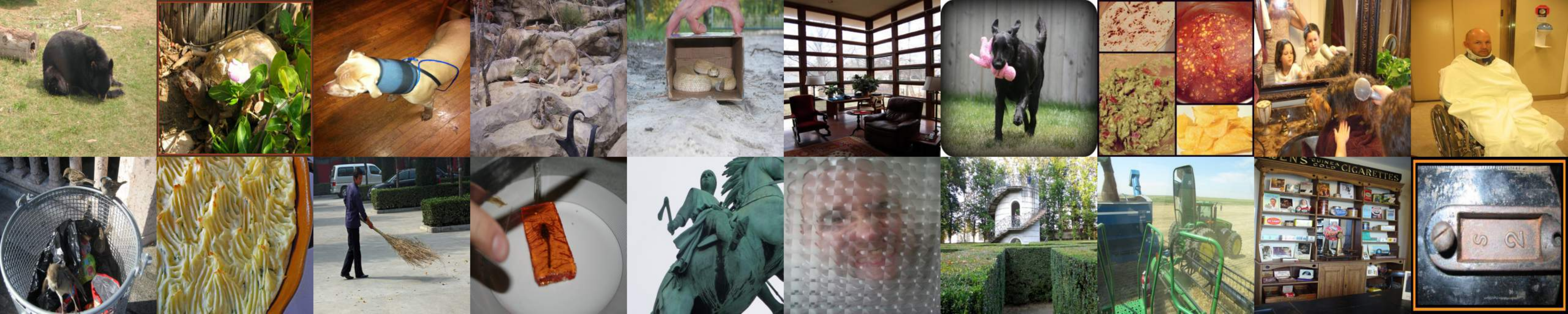}
\caption{Low confidence examples on ImageNet-50.}
\label{fig: low_confidence_imagenet50}
\end{figure}

Figure~\ref{fig: low_confidence_imagenet50} shows examples for which the model produces low confidence predictions on the ImageNet-50 subset. In a number of cases, the low confidence output can be attributed to multiple objects being visible in the scene. Other cases can be explained by the partial visibility of the object, distracting elements in the scene, or ambiguity of the object of interest. 

\subsection{ImageNet - Full}
We include additional qualitative results on the full ImageNet dataset. In particular, Figures~\ref{fig: clusters_imagenet1},~\ref{fig: clusters_imagenet2} and~\ref{fig: clusters_imagenet3} show images from the validation set that were assigned to the same cluster. These can be viewed together with Figure 11 in the main paper. Additionally, we show some mistakes in Figure~\ref{fig: incorrect_clusters}. The failure cases occur when the model focuses too much on the background, or when the network cannot easily discriminate between pairs of similarly looking images. However, in most cases, we can still attach some semantic meaning to the clusters, e.g. animals in cages, white fences.

\begin{figure}
\centering
\includegraphics[width=\textwidth]{ECCV2020/PdfFigures/examples1.pdf}
\caption{Example clusters of ImageNet-1000 (1).}
\label{fig: clusters_imagenet1}
\end{figure}
\begin{figure}
\centering
\includegraphics[width=\textwidth]{ECCV2020/PdfFigures/examples2.pdf}
\caption{Example clusters of ImageNet-1000 (2).}
\label{fig: clusters_imagenet2}
\end{figure}
\begin{figure}
\centering
\includegraphics[width=\textwidth]{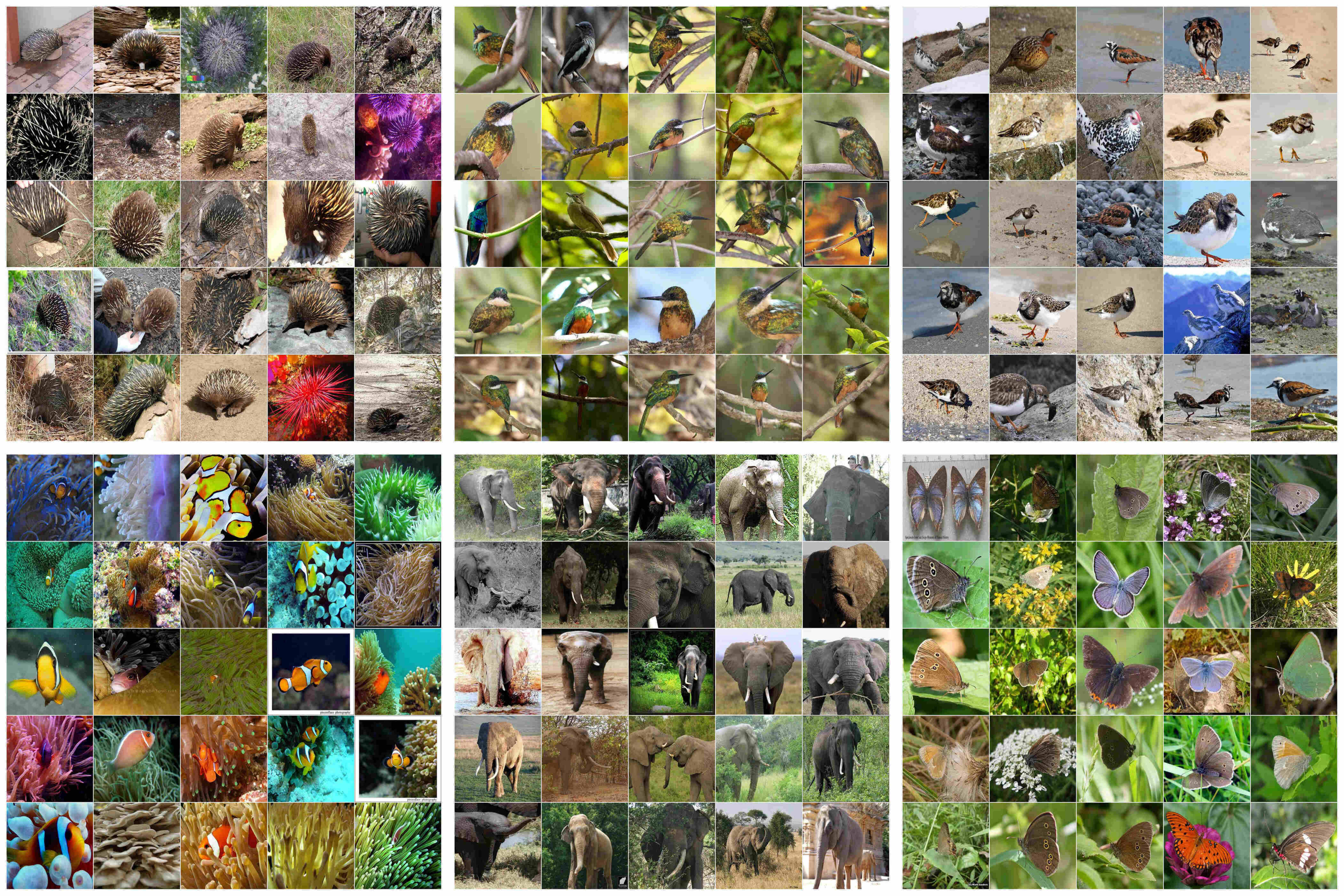}
\caption{Example clusters of ImageNet-1000 (3).}
\label{fig: clusters_imagenet3}
\end{figure}
\begin{figure}
\includegraphics[width=\textwidth]{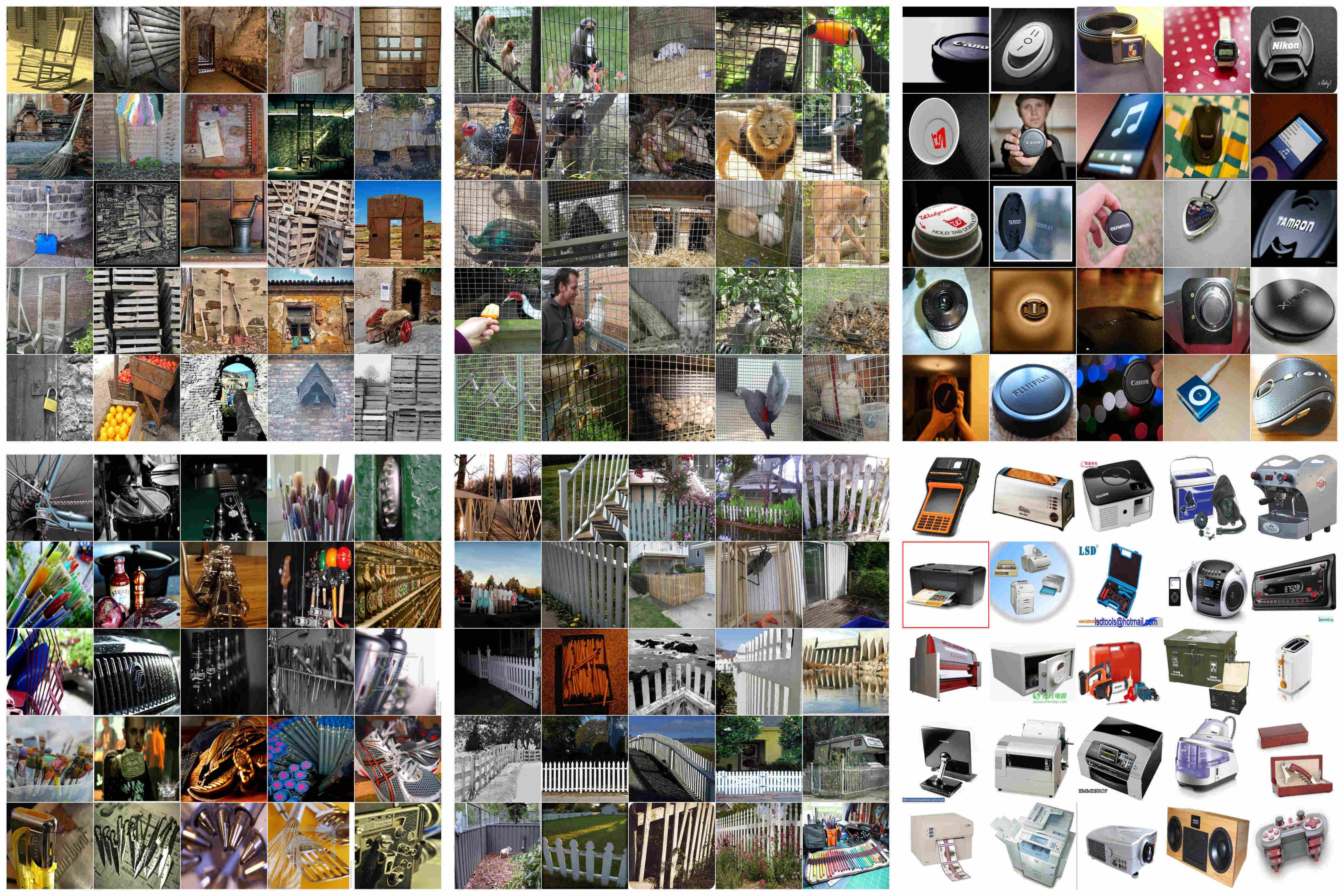}
\caption{Incorrect clusters of ImageNet-1000 predicted by our model.}
\label{fig: incorrect_clusters}
\end{figure}

\section{Experimental setup}
\subsection{Datasets}
\label{subsec:datasets}
Different from prior work~\cite{IIC,DAC,DEC,yang2016joint}, we do not train and evaluate on the full datasets. Differently, we use the standard train-val splits to study the generalization properties of our models. Additionally, we report the mean and standard deviation on the smaller datasets. We would like to encourage future works to adopt this procedure as well. Table~\ref{tab: datasets} provides an overview of the number of classes, the number of images and the aspect ratio of the used datasets. The selected classes on ImageNet-50, ImageNet-100 and ImageNet-200 can be found in our git repository. 

\setlength{\tabcolsep}{4pt}
\begin{table}[ht!]
\scriptsize
\begin{center}
\caption{Datasets overview}
\label{tab: datasets}
\begin{tabular}{@{}l cccc @{}}
\toprule
\textbf{Dataset} & \textbf{Classes} & \textbf{Train images} & \textbf{Val images} & \textbf{Aspect ratio}\\ 
\midrule
CIFAR10 & 10 & 50,000 & 10,000 & 32 x 32\\
CIFAR100-20 & 20 & 50,000 & 10,000 & 32 x 32\\
STL10 & 10 & 5,000 & 8,000 & 96 x 96\\
ImageNet-50 & 50 & 64,274 & 2,500 & 224 x 224\\
ImageNet-100 & 100 & 128,545 & 5,000 & 224 x 224\\
ImageNet-200 & 200 & 256,558 & 10,000 & 224 x 224\\
ImageNet & 1000 & 1,281,167 & 50,000 & 224 x 224\\
\bottomrule
\end{tabular}
\end{center}
\end{table}
\setlength{\tabcolsep}{4pt}
\begin{table}[ht]
\scriptsize
\begin{center}
\caption{List of transformations. The strong transformations are composed by randomly selecting four transformations from the list, followed by Cutout.}
\label{tab: rand_augment}
\begin{tabular}{@{}l cl @{}}
\toprule
\textbf{Transformation} & \textbf{Parameter} & \textbf{Interval}\\ 
\midrule
Identity & - & - \\
Autocontrast & - & - \\
Equalize & - & - \\
Rotate & $\theta$ & $\left[-30,30\right]$ \\
Solarize & $T$ & $\left[0,256\right]$ \\
Color & $C$ & $\left[0.05,0.95\right]$ \\
Contrast & $C$ & $\left[0.05,0.95\right]$ \\
Brightness & $B$ & $\left[0.05,0.95\right]$ \\
Sharpness & $S$ & $\left[0.05,0.95\right]$ \\
Shear X & $R$ & $\left[-0.1,0.1\right]$ \\
Translation X & $\lambda$ & $\left[-0.1,0.1\right]$ \\
Translation Y & $\lambda$ & $\left[-0.1,0.1\right]$ \\
Posterize & $B$ & $\left[4,8\right]$ \\
Shear Y & $R$ & $\left[-0.1,0.1\right]$ \\
\bottomrule
\end{tabular}
\end{center}
\end{table}
\setlength{\tabcolsep}{4pt}

\subsection{Augmentations}
As shown in our experiments, it is beneficial to apply strong augmentations during training. The strong augmentations were composed of four randomly selected transformations from RandAugment~\cite{cubuk2020randaugment}, followed by Cutout~\cite{devries2017improved}. The transformation parameters were uniformly sampled between fixed intervals. Table~\ref{tab: rand_augment} provides a detailed overview. We applied an identical augmentation strategy across all datasets. 

\newpage
\section{Change Log}
The following changes were made since version 1:
\begin{itemize}
    \item Sections 1 and 2: Minor changes were made to the text. References were added to recent related works (CMC, SimCLR, MoCo, FixMatch). Fig. 2 was updated using more recent implementations of the instance discrimination pretext task. 
    \item Section 3.1.: The experimental setup was updated. In particular, we use the implementations from SimCLR and MoCo to perform the instance discrimination task. An identical entropy weight is now used across all datasets. We train and evaluate on the train, val split respectively, rather than using the complete dataset for both training and testing as in prior work. Doing so allows to compare the results against semi- and fully-supervised methods.
    \item Section 3.2.: All results were updated using the SimCLR implementation of the instance discrimination pretext task. Additional experiments were included to study the influence of applying various augmentation strategies, and to analyze the effect of the threshold value.
    \item Section 3.5.: We added a comparison with semi- and fully-supervised methods on ImageNet. We revised our earlier results on the smaller ImageNet subsets due to a coding mistake. Importantly, the initial conclusions remain valid, while results on the full ImageNet dataset improved. We apologize for any inconvenience this might have caused. 
\end{itemize}
\end{document}